\documentclass[letterpaper]{article} 

\usepackage[table]{xcolor}

\usepackage{aaai24}  
\usepackage{times}  
\usepackage{helvet}  
\usepackage{courier}  
\usepackage[hyphens]{url}  
\usepackage{graphicx} 
\usepackage{natbib}  
\usepackage{caption} 
\usepackage{algorithm}
\usepackage{algorithmicx}
\usepackage[noend]{algpseudocode}

\usepackage{newfloat}
\usepackage{listings}
\DeclareCaptionStyle{ruled}{labelfont=normalfont,labelsep=colon,strut=off} 
\floatstyle{ruled}
\newfloat{listing}{tb}{lst}{}
\floatname{listing}{Listing}
\usepackage{float}
\usepackage{tikz}
\usepackage{tikzsymbols}
\usepackage[saturated]{tikzpeople}
\usepackage{mathtools}
\DeclarePairedDelimiter{\abs}{\lvert}{\rvert}

\DeclarePairedDelimiter{\floor}{\lfloor}{\rfloor}
\usetikzlibrary{arrows.meta}
\usepackage{subcaption}
\usepackage{apxproof}
\usepackage{soul} 
\usepackage{amsmath}
\usepackage{tablefootnote}
\usepackage{tabularx}
\let\emph\relax 
\DeclareTextFontCommand{\emph}{\em}
\usepackage{enumitem}
\usepackage{bm}
\usepackage[capitalise]{cleveref}

\newtheorem{theorem}{Theorem}[section]
\newtheorem{corollary}{Corollary}[theorem]

\newtheorem{proposition}[theorem]{Proposition}

\usepackage[utf8]{inputenc}
\usepackage{pgfplots}
\DeclareUnicodeCharacter{2212}{−}
\usepgfplotslibrary{groupplots,dateplot}
\usetikzlibrary{patterns,shapes.arrows}
\pgfplotsset{compat=newest}

\usepackage[utf8]{inputenc} 
\usepackage[T1]{fontenc}    
\usepackage{url}            
\usepackage{booktabs}       
\usepackage{amsfonts}       
\usepackage{nicefrac}       
\usepackage{microtype}      

\usepackage{etoolbox}
\newtoggle{arxiv}
\togglefalse{arxiv}
\toggletrue{arxiv}

\title{
Provably Powerful Graph Neural Networks for Directed Multigraphs 
}

\author{
    B\'{e}ni Egressy\textsuperscript{\rm 1}\thanks{This work was performed while B\'{e}ni Egressy and Luc von Niederhäusern were at IBM Research Europe, Zurich, Switzerland.},
    Luc von Niederhäusern\textsuperscript{\rm 1},
    Jovan Blanu\v{s}a\textsuperscript{\rm 2},
    Erik Altman\textsuperscript{\rm 3},
    Roger Wattenhofer\textsuperscript{\rm 1},
    Kubilay Atasu\textsuperscript{\rm 2}
}
\affiliations{
    \textsuperscript{\rm 1}ETH Zurich, Zurich, Switzerland\\
    \textsuperscript{\rm 2}IBM Research Europe, Zurich, Switzerland\\
    \textsuperscript{\rm 3}IBM Watson Research, Yorktown Heights, NY, USA\\
    begressy@ethz.ch, lucv@ethz.ch, jov@zurich.ibm.com, ealtman@us.ibm.com, wattenhofer@ethz.ch, kat@zurich.ibm.com
}

\begin{document}

\maketitle

\begin{abstract}
This paper analyses a set of simple adaptations that transform standard message-passing Graph Neural Networks (GNN) into provably powerful directed multigraph neural networks. The adaptations include multigraph port numbering, ego IDs, and reverse message passing. We prove that the combination of these theoretically enables the detection of any directed subgraph pattern. To validate the effectiveness of our proposed adaptations in practice, we conduct experiments on synthetic subgraph detection tasks, which demonstrate outstanding performance with almost perfect results.

Moreover, we apply our proposed adaptations to two financial crime analysis tasks. We observe dramatic improvements in detecting money laundering transactions, improving the minority-class F1 score of a standard message-passing GNN by up to $30\%$, and closely matching or outperforming tree-based and GNN baselines. Similarly impressive results are observed on a real-world phishing detection dataset, boosting three standard GNNs' F1 scores by around $15\%$ and outperforming all baselines. 
\iftoggle{arxiv}{%
    To appear as a conference paper at AAAI 2024.
}{%
    An extended version with appendices can be found on arXiv: \url{https://arxiv.org/abs/2306.11586}.
}
\end{abstract}

\section{Introduction}
\label{sec:intro}



Graph neural networks (GNNs) have become the go-to machine learning models for learning from relational data. GNNs are used in various fields, ranging from biology, physics, and chemistry to social networks, traffic, and weather forecasting \citep{bongini2021molecular, zhou2020graph_survey, derrow2021eta, shu2019beyondFakeNews, wu2020comprehensiveSurvey, keisler2022forecastingWeather, zhang2019deep_recommend, battaglia2016interaction_physics}. More recently, there has been growing interest in using GNNs to identify financial crime \citep{cardoso2022laundrograph, kanezashi2022ethereum, weber2019anti, weber2018scalable, nicholls2021financial}.

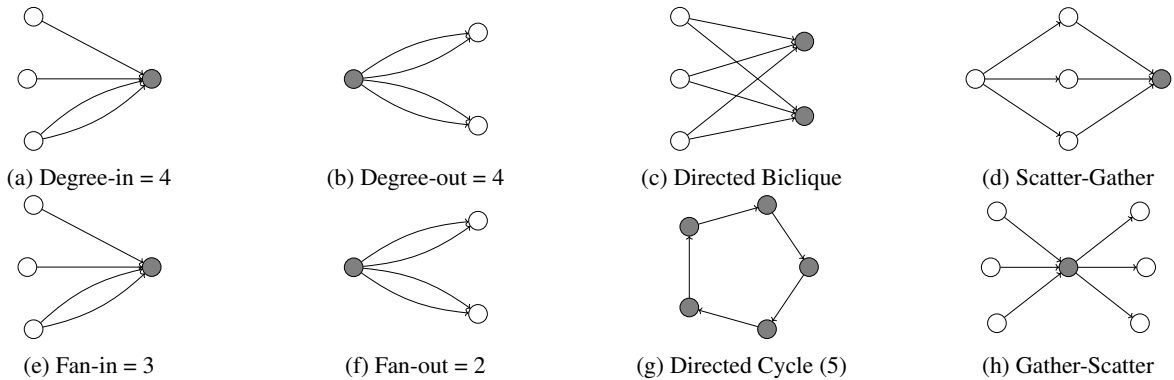
\begin{figure*}
\centering
\begin{subfigure}{.21\textwidth}
\centering
\resizebox{0.75\linewidth}{!}{
\begin{tikzpicture}
    \node[circle, minimum size=0.6cm] (O1) at  (-3,-2) {};
    \node[circle, minimum size=0.6cm] (O2) at  (3,2) {};
    \node[circle, draw, minimum size=0.6cm] (B1) at  (-1.8,-2) {};
    \node[circle, draw, minimum size=0.6cm] (B3) at  (-2,0) {};
    \node[circle, draw, minimum size=0.6cm] (B5) at  (-1.8,2) {};
    \node[circle, draw, minimum size=0.6cm, fill=gray] (C) at  (2,0) {};
    \draw [semithick,->] (B1) edge [bend left=15] (C);
    \draw [semithick,->] (B1) edge [bend right=15] (C);
    \draw [semithick,->] (B3) -- (C);
    \draw [semithick,->] (B5) -- (C);
\end{tikzpicture}
}
\caption{Degree-in = 4}\label{subfig:deg_in}
\end{subfigure}
\hfil
\begin{subfigure}{.21\textwidth}
\centering
\resizebox{0.75\linewidth}{!}{
\begin{tikzpicture}
    \node[circle, minimum size=0.6cm] (O1) at  (-3,-2) {};
    \node[circle, minimum size=0.6cm] (O2) at  (3,2) {};
    \node[circle, draw, minimum size=0.6cm, fill=gray] (A) at  (-2,0) {};
    \node[circle, draw, minimum size=0.6cm] (B1) at  (2,-1.5) {};
    \node[circle, draw, minimum size=0.6cm] (B2) at  (2,1.5) {};
    \draw [semithick,->] (A) edge [bend left=15] (B1);
    \draw [semithick,->] (A) edge [bend right=15] (B1);
    \draw [semithick,->] (A) edge [bend left=15] (B2);
    \draw [semithick,->] (A) edge [bend right=15] (B2);
\end{tikzpicture}
}
\caption{Degree-out = 4}\label{subfig:deg_out}
\end{subfigure}
\hfil
\begin{subfigure}{.21\textwidth}
\centering
\resizebox{0.75\linewidth}{!}{
\begin{tikzpicture}
    \node[circle, minimum size=0.6cm] (O1) at  (-3,-2) {};
    \node[circle, minimum size=0.6cm] (O2) at  (3,2) {};
    \node[circle, draw, minimum size=0.6cm] (B1) at  (-2,-2) {};
    \node[circle, draw, minimum size=0.6cm] (B2) at  (-2,0) {};
    \node[circle, draw, minimum size=0.6cm] (B3) at  (-2,2) {};
    \node[circle, draw, minimum size=0.6cm, fill=gray] (C1) at  (2,1.2) {};
    \node[circle, draw, minimum size=0.6cm, fill=gray] (C2) at  (2,-1.2) {};
    \draw [semithick,->] (B1) edge (C1);
    \draw [semithick,->] (B2) edge (C1);
    \draw [semithick,->] (B3) edge (C1);
    \draw [semithick,->] (B1) edge (C2);
    \draw [semithick,->] (B2) edge (C2);
    \draw [semithick,->] (B3) edge (C2);
\end{tikzpicture}
}
\caption{Directed Biclique}\label{subfig:biclique}
\end{subfigure}
\hfil
\begin{subfigure}{.21\textwidth}
\centering
\resizebox{0.75\linewidth}{!}{
\begin{tikzpicture}
    \node[circle, minimum size=0.6cm] (O1) at  (-3,-2) {};
    \node[circle, minimum size=0.6cm] (O2) at  (3,2) {};
    \node[circle, draw, minimum size=0.6cm] (A) at  (-3,0) {};
    \node[circle, draw, minimum size=0.6cm] (B1) at  (0,-2) {};
    \node[circle, draw, minimum size=0.6cm] (B2) at  (0,0) {};
    \node[circle, draw, minimum size=0.6cm] (B3) at  (0,2) {};
    \node[circle, draw, minimum size=0.6cm, fill=gray] (C) at  (3,0) {};
    \draw [semithick,->] (A) -- (B1);
    \draw [semithick,->] (A) -- (B2);
    \draw [semithick,->] (A) -- (B3);
    \draw [semithick,->] (B1) -- (C);
    \draw [semithick,->] (B2) -- (C);
    \draw [semithick,->] (B3) -- (C);
\end{tikzpicture}
}
\caption{Scatter-Gather}\label{subfig:scatter-gather}
\end{subfigure}
\vfil
\begin{subfigure}{.21\textwidth}
\centering
\resizebox{0.75\linewidth}{!}{
\begin{tikzpicture}
    \node[circle, minimum size=0.6cm] (O1) at  (-3,-2) {};
    \node[circle, minimum size=0.6cm] (O2) at  (3,2) {};
    \node[circle, draw, minimum size=0.6cm] (B1) at  (-1.8,-2) {};
    \node[circle, draw, minimum size=0.6cm] (B3) at  (-2,0) {};
    \node[circle, draw, minimum size=0.6cm] (B5) at  (-1.8,2) {};
    \node[circle, draw, minimum size=0.6cm, fill=gray] (C) at  (2,0) {};
    \draw [semithick,->] (B1) edge [bend left=15] (C);
    \draw [semithick,->] (B1) edge [bend right=15] (C);
    \draw [semithick,->] (B3) -- (C);
    \draw [semithick,->] (B5) -- (C);
\end{tikzpicture}
}
\caption{Fan-in = 3}\label{subfig:fan_in}
\end{subfigure}
\hfil
\begin{subfigure}{.21\textwidth}
\centering
\resizebox{0.75\linewidth}{!}{
\begin{tikzpicture}
    \node[circle, minimum size=0.6cm] (O1) at  (-3,-2) {};
    \node[circle, minimum size=0.6cm] (O2) at  (3,2) {};
    \node[circle, draw, minimum size=0.6cm, fill=gray] (A) at  (-2,0) {};
    \node[circle, draw, minimum size=0.6cm] (B1) at  (2,-1.5) {};
    \node[circle, draw, minimum size=0.6cm] (B2) at  (2,1.5) {};
    \draw [semithick,->] (A) edge [bend left=15] (B1);
    \draw [semithick,->] (A) edge [bend right=15] (B1);
    \draw [semithick,->] (A) edge [bend left=15] (B2);
    \draw [semithick,->] (A) edge [bend right=15] (B2);
\end{tikzpicture}
}
\caption{Fan-out = 2}\label{subfig:fan_out}
\end{subfigure}
\hfil
\begin{subfigure}{.21\textwidth}
\centering
\resizebox{0.75\linewidth}{!}{
\begin{tikzpicture}
    \node[circle, minimum size=0.6cm] (O1) at  (-3,-2) {};
    \node[circle, minimum size=0.6cm] (O2) at  (3,2) {};
    \node[circle, draw, minimum size=0.6cm, fill=gray] (C1) at  (-1.7,1.3) {};
    \node[circle, draw, minimum size=0.6cm, fill=gray] (C2) at  (0.8,2) {};
    \node[circle, draw, minimum size=0.6cm, fill=gray] (C3) at  (2.15,0) {};
    \node[circle, draw, minimum size=0.6cm, fill=gray] (C4) at  (0.8,-2) {};
    \node[circle, draw, minimum size=0.6cm, fill=gray] (C5) at  (-1.7,-1.3) {};
    \draw [semithick,->] (C1) -- (C2);
    \draw [semithick,->] (C2) -- (C3);
    \draw [semithick,->] (C3) -- (C4);
    \draw [semithick,->] (C4) -- (C5);
    \draw [semithick,->] (C5) -- (C1);
\end{tikzpicture}
}
\caption{Directed Cycle (5)}\label{subfig:cycle}
\end{subfigure}
\hfil
\begin{subfigure}{.21\textwidth}
\centering
\resizebox{0.75\linewidth}{!}{
\begin{tikzpicture}
    \node[circle, minimum size=0.6cm] (O1) at  (-3,-2) {};
    \node[circle, minimum size=0.6cm] (O2) at  (3,2) {};
    \node[circle, draw, minimum size=0.6cm] (A1) at  (-2.3,-1.8) {};
    \node[circle, draw, minimum size=0.6cm] (A2) at  (-2.5,0) {};
    \node[circle, draw, minimum size=0.6cm] (A3) at  (-2.3,1.8) {};
    \node[circle, draw, minimum size=0.6cm] (C1) at  (2.3,-1.8) {};
    \node[circle, draw, minimum size=0.6cm] (C2) at  (2.5,0) {};
    \node[circle, draw, minimum size=0.6cm] (C3) at  (2.3,1.8) {};
    \node[circle, draw, minimum size=0.6cm, fill=gray] (B) at  (0,0) {};
    \draw [semithick,->] (A1) -- (B);
    \draw [semithick,->] (A2) -- (B);
    \draw [semithick,->] (A3) -- (B);
    \draw [semithick,->] (B) -- (C1);
    \draw [semithick,->] (B) -- (C2);
    \draw [semithick,->] (B) -- (C3);
\end{tikzpicture}
}
\caption{Gather-Scatter}\label{subfig:gather-scatter}
\end{subfigure}
\caption{Money Laundering Patterns. The gray fill indicates the nodes to be detected by the synthetic pattern detection tasks. The exact degree/fan pattern sizes here are for illustrative purposes only. 
}
\label{fig:aml_patterns}
\end{figure*}

Our motivating task is to detect financial crimes manifesting as subgraph patterns in transaction networks. For example, see \cref{fig:aml_patterns}, which depicts established money laundering 
\iftoggle{arxiv}{%
    patterns, or see \cref{fig:intro_ex} in the appendix for an illustrative scenario.
}{%
    patterns.
}
But note that similar patterns are relevant for graph tasks in many areas, ranging from chemistry to traffic forecasting.
The task seems to lend itself nicely to the use of GNNs. Unfortunately,
current GNNs are ill-equipped to deal with financial transaction networks effectively. 

Firstly, financial transaction networks are, in fact, directed multigraphs, i.e., edges (or transactions) have a direction, and there can be multiple edges between two nodes (or accounts). 
Secondly, most GNNs cannot detect some subgraph patterns, such as cycles \citep{chen2020canSubgraphCounting, chen2019equivalence}. There have been many efforts to overcome this limitation \citep{you2021identity, huang2022boostingCycleCount, papp2022theoretical, zhang2021nested, loukas2019graph, sato2019approximation_port_numbers}, all focusing on simple (undirected) graphs. But even on simple graphs, the problem is far from solved. Until very recently, for example, there was no linear-time permutation-equivariant GNN that could count 6-cycles with theoretical guarantees \citep{huang2022boostingCycleCount}.

This paper addresses both of these issues. To our knowledge, this is the first GNN architecture designed specifically for directed multigraphs. Secondly, we first prove that the proposed architecture can theoretically detect any subgraph pattern in directed multigraphs and then empirically confirm that our proposed architecture can detect the patterns illustrated in \cref{fig:aml_patterns}.
Our proposed architecture is based on a set of simple adaptations that can transform any standard GNN architecture into a directed multigraph GNN. The adaptations are reverse message passing \citep{jaume2019edgnnbidirecibmmaria}, port numbering \citep{sato2019approximation_port_numbers}, and ego IDs \citep{you2021identity}.
Although these individual building blocks are present in existing literature, the theoretical and empirical power of combining them has not been explored. In this work, we fill this gap: We combine them, adapt them to directed multigraphs, and showcase the theoretical and empirical advantages of using them in unison.

\textbf{Our contributions.}
(1) We propose a set of simple and intuitive adaptations that can transform message-passing GNNs into provably powerful directed multigraph neural networks. 
(2) We prove that suitably powerful GNNs equipped with ego IDs, port numbering, and reverse message passing can identify any directed subgraph pattern.
(3) The theory is tested on synthetic graphs, confirming that GNNs using these adaptations can detect a variety of subgraph patterns, including directed cycles up to length six, scatter-gather patterns, and directed bicliques, setting them apart from previous GNN architectures.
(4) The improvements translate to significant gains on two financial datasets. The adaptations boost GNN performance dramatically on money laundering and phishing datasets, matching or surpassing state-of-the-art financial crime detection models on both simulated and real data.

\section{Related Work}
\label{sec:related}

\citet{xu2018powerful} showed that standard MPNNs are at most as powerful as the Weisfeiler-Lehman (WL) isomorphism test, and provided a GNN architecture, GIN, that theoretically matches the power of the WL test. 
Although the WL test can asymptotically almost surely differentiate any two non-isomorphic graphs \citep{babai1980random}, standard MPNNs cannot --- in certain graphs --- detect simple substructures like cycles \citep{chen2020canSubgraphCounting, chen2019equivalence}.  
This motivated researchers to go beyond standard MPNNs. 

One direction considers emulating the more powerful k-WL isomorphism test, by conducting message passing between k-tuples or using a tensor-based model \citep{maron2019provably, morris2019weisfeiler}. Unfortunately, these models have high complexity and are impractical for most applications.
Another line of work uses pre-calculated features to augment the GNN. These works explore adding subgraph counts \citep{bouritsas2022improving, barcelo2021graph}, positional node embeddings \citep{egressy2022graph, dwivedi2021graph}, random IDs \citep{abboud2020randomIDs, sato2021random}, and node IDs \citep{loukas2019graph}. 

A recent class of expressive GNNs called Subgraph GNNs, model graphs as collections of subgraphs \citep{frasca2022understanding, zhao2021stars}. \citet{papp2021dropgnn} drop random nodes from the input and run the GNN multiple times, gathering more information with each run. \citet{zhang2021nested} instead extract subgraphs around each node and run the GNN on these. 
Also falling into this category is ID-GNN, which uses ego IDs \citep{you2021identity}, whereby each node is sampled with its neighborhood and given an identifier to differentiate it from the neighbors. Although the authors claim that ID-GNNs can count cycles, the proof turns out to be incorrect. In fact, \citet{huang2022boostingCycleCount} show that the whole family of Subgraph GNNs cannot count cycles of length greater than 4, and propose I$^2$-GNNs that can count cycles up to length 6.

There has been much less work on GNNs for directed graphs. \citet{zhang2021magnet} propose a spectral network for directed graphs, but it is difficult to analyze the power of this network or apply it to larger datasets. Similar approaches can be found in \citep{tong2020digraph} and \citep{ma2019spectral}. 
\citet{jaume2019edgnnbidirecibmmaria} extend message passing to aggregate incoming and outgoing neighbors separately, rather than naively treating the graph as undirected. 
Directed multigraphs have not specifically been considered.

GNNs have been used for various financial applications \citep{li2021stockModeling, feng2019temporalstock, chen2018incorporatingstocks, zhang2019deep_recommend, li2019hierarchicalrecommend, xu2021towardsdefault, yang2021financialdefault}. 
Closest to our work, GNNs have been used for fraud detection. \citet{liang2019uncoveringfraud} and \citet{rao2021xfraud} work on bipartite customer-product graphs to uncover insurance and credit card fraud, respectively. \citet{liu2018heterogeneous} use heterogeneous GNNs to detect malicious accounts in the device-activity bipartite graph of an online payment platform. 
\citet{weber2019anti} were the first to apply standard GNNs for anti-money laundering (AML), and more recently \citet{cardoso2022laundrograph} proposed representing the transaction network as a bipartite account-transaction graph and showed promising results in the semi-supervised AML setting. However, it is not clear how these approaches help with detecting typical fraud patterns. 





\section{Background}
\label{sec:background}

\subsection{Graphs and Financial Transaction Graphs}

We consider directed multigraphs, $G$, where the nodes $v \in V(G)$ represent accounts, and the directed edges $e=(u,v) \in E(G)$ represent transactions from $u$ to $v$. Each node $u$ (optionally) has a set of account features $h^{(0)}(u)$; this could include the account number, bank ID, and account balance. Each transaction $e=(u,v)$ has a set of associated transaction features $h^{(0)}_{(u,v)}$; this includes the amount, currency, and timestamp of the transaction. 
The incoming and outgoing neighbors of $u$ are denoted by $N_{in}(u)$ and $N_{out}(u)$ respectively. Multiple transactions between the same two accounts are possible, making $G$ a multigraph. In node (or edge) prediction tasks, each node (or edge) will have a binary label indicating whether the account (or transaction) is illicit. 


\noindent
\textbf{Financial Crime Patterns.} \hspace{0.2cm} 
\cref{fig:aml_patterns} shows a selection of subgraph patterns indicative of money laundering \citep{granados2022AMLgeometry, he2021efficientAMLpatterns, amlsim, weber2018scalable, starnini2021smurf_italianbank}.
Unfortunately, these are rather generic patterns, which also appear extensively amongst perfectly innocent transactions. 
As a result, detecting financial crime relies not just on detecting individual patterns, but also on learning relevant combinations. This makes neural networks promising candidates for the task. However, standard message-passing GNNs typically fail to detect the depicted patterns, except for degree-in. 
In the next section, we describe architectural adaptations, which enable GNNs to detect each one of these patterns.

\noindent
\textbf{Subgraph Detection.} \hspace{0.2cm} 
Given a subgraph pattern $H$, we define subgraph detection for nodes as deciding for each node in a graph whether it is part of a subgraph that is isomorphic to $H$; i.e., given a node $v \in V(G)$, deciding whether there exists a graph $G'$, with $E(G') \subseteq E(G)$ and $V(G') \subseteq V(G)$, such that $v \in V(G')$ and $G' \cong H$.

\subsection{Message Passing Neural Networks}

Message-passing GNNs, commonly referred to as Message Passing Neural Networks (MPNNs), form the most prominent family of GNNs. They include GCN \citep{kipf2016semiGCN}, GIN \citep{xu2018powerful}, GAT \citep{velivckovic2017graph}, GraphSAGE \citep{hamilton2017inductive}, and many more architectures. They work in three steps: (1) Each node sends a message with its current state $h(v)$ to its neighbors, (2) Each node aggregates all the messages it receives from its neighbors in the embedding $a(v)$, and (3) Each node updates its state based on $h(v)$ and $a(v)$ to produce a new state. These 3 steps constitute a layer of the GNN, and they can be repeated to gather information from further and further reaches of the graph. More formally:
\begin{align*}
    a^{(t)}(v) &= \textsc{Aggregate} \left( \{ h^{(t-1)}(u) \mid u \in N(v) \} \right), \\
    h^{(t)}(v) &= \textsc{Update} \left( h^{(t-1)}(v), a^{(t)}(v) \right),
\end{align*}
where $\{\{ . \}\}$ denotes a multiset, and \textsc{Aggregate} is a permutation-invariant function. We will shorten $\textsc{Aggregate}$ to $\textsc{Agg}$, and for readability, we will use $\{ . \}$ rather than $\{\{ . \}\}$ to indicate multisets.

In the case of directed graphs, we need to distinguish between the incoming and outgoing neighbors of node $u$. In a standard MPNN, the messages are passed along the directed edges in the direction indicated. As such, the aggregation step only considers messages from incoming neighbors:
\begin{align*}
    a^{(t)}(v) &= \textsc{Agg} \left( \{ h^{(t-1)}(u) \mid u \in N_{in}(v) \} \right),
\end{align*}
where we aggregate over the incoming neighbors, $N_{in}(v)$.

The edges of an input graph may also have input features. We denote the input features of directed edge $e=(u,v)$ by $h^{(0)}{\left((u,v)\right)}$. When using edge features during the message passing, the aggregation step becomes:
\begin{align*}
    a^{(t)}(v) &= \textsc{Agg} \left( \{ ( h^{(t-1)}(u), h^{(0)}{\left((u,v)\right)} ) \mid u \in N_{in}(v) \} \right)
\end{align*}
In the remainder, we omit edge features from formulas when unnecessary in favor of brevity.

\section{Methods}
\label{sec:methods}

In this section, we introduce simple adaptations for standard MPNNs (Message Passing Neural Networks) to enable the detection of the fraud patterns in \cref{fig:aml_patterns}. We consider the adaptations in increasing order of complexity in terms of the patterns they help to detect. 
We provide theory results to motivate the adaptations and include corresponding experiments on the synthetic subgraph detection dataset in \cref{subsec:subgraph_pattern_results} to support the theory empirically.

\subsection{Reverse Message Passing}

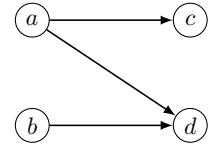
\begin{figure}
    \centering
    \begin{tikzpicture}[scale=0.7, every node/.style={scale=0.9}, >=latex]
        \node (O1) at  (-1.5,2.4) {};
        \node (O2) at  (1.5,-0.4) {};
        
        \node[circle, draw, minimum size=0.5cm, inner sep=0] (A) at  (-1.5,2) {$a$};
        \node[circle, draw, minimum size=0.5cm, inner sep=0] (B) at  (-1.5,0) {$b$};
        \node[circle, draw, minimum size=0.5cm, inner sep=0] (C) at  (1.5,2) {$c$};
        \node[circle, draw, minimum size=0.5cm, inner sep=0] (D) at  (1.5,0) {$d$};
        \draw [semithick,->] (A) -- (C);
        \draw [semithick,->] (A) -- (D);
        \draw [semithick,->] (B) -- (D);
    \end{tikzpicture}
    \caption{Nodes ($a$ and $b$) with different out-degrees are not distinguishable by a standard MPNN with directed message passing. Note that naive bidirectional message passing, on the other hand, is unable to distinguish nodes $a$ and $d$.
    }
    \label{fig:deg_out_ex}
\end{figure}

When using a standard MPNN with directed edges, a node does not receive any messages from outgoing neighbors (unless they happen to also be incoming neighbors), and so is unable to count its outgoing edges. For example, a standard MPNN is unable to distinguish nodes $a$ and $b$ in \cref{fig:deg_out_ex}. 
Further, note that naive bidirectional message passing, where edges are treated as undirected and messages travel in both directions, does not solve the problem, because a node then can not distinguish incoming and outgoing edges. So this would fail to distinguish nodes $a$ and $d$ in the same figure. 

To overcome this issue, we need to indicate the direction of the edges in some way. We propose using a separate message-passing layer for the incoming and outgoing edges respectively, i.e., adding \emph{reverse message passing}. Note that this is akin to using a relational GNN with two edge types \citep{schlichtkrull2018modelingRelGNN}. 
More formally, the aggregation and update mechanisms become:
\begin{align*}
    a_{in}^{(t)}(v) &= \textsc{Agg}_{in} \left( \{ h^{(t-1)}(u) \mid u \in N_{in}(v) \} \right), \\
    a_{out}^{(t)}(v) &= \textsc{Agg}_{out} \left( \{ h^{(t-1)}(u) \mid u \in N_{out}(v) \} \right), \\
    h^{(t)}(v) &= \textsc{Update} \left( h^{(t-1)}(v), a_{in}^{(t)}(v), a_{out}^{(t)}(v) \right) ,
\end{align*}
where $a_{in}$ is now an aggregation of incoming neighbors and $a_{out}$ of outgoing neighbors. We now prove that message-passing GNNs with reverse MP can solve degree-out.


\begin{proposition}
\label{thm:reverseMP_outdegree}
An MPNN with sum aggregation and reverse MP can solve degree-out.
\end{proposition}



The proof of \cref{thm:reverseMP_outdegree} can be found 
\iftoggle{arxiv}{%
    in \cref{app:proofs}.
}{%
    in the appendix.
}
In \cref{subsec:subgraph_pattern_results}, we use a synthetic pattern detection task to confirm that the theory translates into practice. 

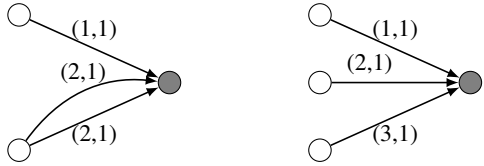
\begin{figure}
    \centering
    \begin{tikzpicture}[scale=0.5, every node/.style={scale=0.9}, >=latex]
        \node[circle, draw, minimum size=0.3cm] (A) at  (0,0) {};
        \node[circle, draw, minimum size=0.3cm] (B) at  (0,3.6) {};
        \node[circle, draw, minimum size=0.3cm, fill=gray] (C) at  (4,1.8) {};
        \draw [semithick,->] (A) edge [bend left] node[pos=0.5,above] {(2,1)} (C);
        \draw [semithick,->] (A) -- (C) node[pos=0.5,below] {(2,1)};
        \draw [semithick,->] (B) -- (C) node[pos=0.5,above] {(1,1)};
        
        \node[circle, draw, minimum size=0.3cm] (A2) at  (8,3.6) {};
        \node[circle, draw, minimum size=0.3cm] (B2) at  (8,1.8) {};
        \node[circle, draw, minimum size=0.3cm, fill=gray] (C2) at  (12,1.8) {};
        \node[circle, draw, minimum size=0.3cm] (D2) at  (8,0) {};
        \draw [semithick,->] (A2) -- (C2) node[pos=0.5,above] {(1,1)};
        \draw [semithick,->] (B2) -- (C2) node[pos=0.3,above] {(2,1)};
        \draw [semithick,->] (D2) -- (C2) node[pos=0.5,below] {(3,1)};
    \end{tikzpicture}
    \caption{Nodes (in gray) with different fan-ins that are not distinguishable by a standard MPNN. The edge labels indicate incoming and outgoing port numbers, respectively.}
    \label{fig:fan_in_ex}
\end{figure}

\subsection{Directed Multigraph Port Numbering}

People often make multiple transactions to the same account. In transaction networks, these are represented as parallel edges. To detect fan-in (or fan-out) patterns, a model has to distinguish between edges from the same neighbor and edges from different neighbors. 
Using unique account numbers (or in general node IDs) would naturally allow for this. However, using account numbers does not generalize well. During training, a model can memorize fraudulent account numbers without learning to identify fraudulent patterns, but this will not generalize to unseen accounts.

Instead, we adapt port numbering \citep{sato2019approximation_port_numbers} to directed multigraphs. Port numbering assigns local IDs to each neighbor at a node. This allows a node to identify messages coming from the same neighbor in consecutive message-passing rounds. 
To adapt port numbering to directed multigraphs, we assign each directed edge an incoming and an outgoing port number, and edges coming from (or going to) the same node, receive the same incoming (or outgoing) port number. 
Unlike \citet{sato2019approximation_port_numbers}, who attach only the local port numbers at a node to received messages, we attach the port numbers in both directions, i.e., a node sees both the port number it has assigned to a neighbor and the port number that the neighbor has assigned to it. 
This turns out to be crucial for our expressivity arguments. 

Port numbers have been shown to increase the expressivity of GNNs on simple graphs, but message-passing GNNs with port numbers alone cannot even detect 3-cycles in some cases \citep{garg2020generalization_jegelka_ports}.

In general, the assignment of port numbers around a node is arbitrary. A node with $d$ incoming neighbors can assign incoming port numbers in $d!$ ways. To break this symmetry in our datasets, we use the transaction timestamps to order the incoming (or outgoing) neighbors. In the case of parallel edges, we use the earliest timestamp to decide the order of the neighbors. Since timestamps carry meaning in financial crime detection, the choice of ordering is motivated; indeed two identical subgraph patterns with different timestamps can have different meanings. 

Computing the port numbers in this way can be a time-intensive step, with runtime complexity dominated by sorting all edges by their timestamps: $\mathcal{O}(m \log m)$, where $m = \abs{E(G)}$. However, port numbers can be calculated in advance, 
so training and inference times are unaffected. 
Each edge receives an incoming and an outgoing port number
as additional edge features.
\cref{fig:fan_in_ex} shows an example of graphs with port numbers.
We now prove that GNNs using port numbers can correctly identify fan-in and fan-out patterns. 

Note that the following proof, and later proofs using port numbers, do not rely on the timestamps for correctness. 
However, if timestamps that uniquely identify the ports are available, then permutation invariance/equivariance of the GNN will be preserved.

\begin{proposition}
\label{thm:ports_fanin}
An MPNN with max aggregation and multigraph port numbering can solve fan-in.
\end{proposition}



A proof is provided 
\iftoggle{arxiv}{%
    in \cref{app:proofs}.
}{%
    in the appendix.
} 
Adding reverse MP, one can argue similarly that fan-out can also be solved. Both propositions are confirmed empirically in \cref{subsec:subgraph_pattern_results}. 

\begin{proposition}
\label{thm:fanout}
An MPNN with max aggregation, multigraph port numbering, and reverse MP can solve fan-out.
\end{proposition}


\subsection{Ego IDs}

Although reverse MP and multigraph port numbering help with detecting some of the suspicious patterns in \cref{fig:aml_patterns}, they are not sufficient to detect directed cycles, scatter-gather patterns, and directed bicliques.  
\citet{you2021identity} introduced ego IDs specifically to help detect cycles in graphs. The idea is that by ``marking'' a ``center'' node with a distinct (binary) feature, this node can recognize when a sequence of messages cycles back around to it, thereby detecting cycles that it is part of. 
However, it turns out that the proof of Proposition~2 in the paper is incorrect, and ego IDs alone do not enable cycle detection. We give a counterexample
\iftoggle{arxiv}{%
    in \cref{fig:ego_fail_ex} in the appendix.
}{%
    in the appendix.
}
Indeed, \citet{huang2022boostingCycleCount} also note that the proof ``confuses walks with paths''. 


We see this reflected in the individual results in \cref{tab:sim_results}. Although ego IDs offer a boost in detecting short cycles, they do not help the baseline (GIN) in detecting longer cycles. 
This can also be explained theoretically: Assuming a graph has no loops (edges from a node to itself), walks of length two and three that return to the start node are also cycles since there is no possibility to repeat intermediate nodes. Therefore Proposition~2 from \citet{you2021identity} applies in these cases and it is not surprising that GIN+EgoIDs can achieve impressive F1 scores for $2$- and $3$-cycle detection.

However, in combination with reverse MP and port numbering, ego IDs can detect cycles, scatter-gather patterns, and bipartite subgraphs, completing the list of suspicious patterns. In fact, it can be shown that a suitably powerful standard MPNN with these adaptations can distinguish any two non-isomorphic (sub-)graphs, and given a \emph{consistent} use of port-numbering they will not mistakenly distinguish any two isomorphic (sub-)graphs. GNNs fulfilling these two properties are often referred to as \emph{universal}. The crux of the proof is showing how the ego ID, port numbers, and reverse MP can be used to assign unique IDs to each node in the graph. Given unique node IDs, sufficiently powerful standard MPNNs are known to be universal \citep{loukas2019graph, abboud2020randomIDs}.

\begin{theorem}
\label{thm:nodeIDs}
Ego IDs combined with port numbering and reverse MP can be used to assign unique node IDs in connected directed multigraphs.
\end{theorem}

The idea of the proof is to show how a GNN can replicate a labeling algorithm that assigns unique IDs to each node in an ego node's neighborhood. The labeling algorithm as well as the full proof are provided 
\iftoggle{arxiv}{%
    in \cref{app:proofs}.
}{%
    in the appendix.
} 
The universality of the adaptations follows from this theorem.

\begin{corollary}
\label{cor:universality}
GIN with ego IDs, port numbering, and reverse MP can theoretically detect any directed subgraph pattern.
\end{corollary}

The proof follows from \cref{thm:nodeIDs} above and Corollary~3.1 from \citet{loukas2019graph}. 
Similar statements can be made for simple undirected graphs. One can remove the reverse MP from the assumptions since this is only needed to make the proof work with directed edges. 

\begin{theorem}
\label{thm:nodeIDs_undirected}
Ego IDs and port numbering can be used to assign unique node IDs in connected undirected graphs.
\end{theorem}

\begin{corollary}
\label{cor:universality_undirected}
GIN with ego IDs and port numbering can theoretically detect any subgraph in undirected graphs.
\end{corollary}

The ablation study in \cref{tab:sim_results} of \cref{subsec:subgraph_pattern_results} again supports the theoretical analysis. 
The combination of the three adaptations achieves impressive scores for all subgraph patterns.

Note that passing the port numbers of both incident nodes of an edge is crucial for inferring unique node IDs. 
\iftoggle{arxiv}{%
    \cref{fig:ex_bidirec_ports} in the appendix illustrates this with a simple example.
}{%
    We illustrate this with a simple example in the appendix.
}
In particular, port numbering, as introduced by \citet{sato2019approximation_port_numbers}, is not sufficient. 

\subsection{Complexity \& Runtime}



We propose a set of adaptations, so the final model complexity will depend on the choice of base GNN.
We describe the additional runtime costs incurred by the adaptations 
\iftoggle{arxiv}{%
    in \cref{app:complexity}.
}{%
    in the appendix.
}
All in all, the adaptations add a constant factor to the runtime complexity
in addition to a one-off pre-computation cost of $\mathcal{O}(m \log(m))$.
The empirical runtimes on AML Small HI using GIN can be seen in 
\iftoggle{arxiv}{%
    \cref{app:runtimes}.
}{%
    in the appendix.
} 


\section{Datasets}
\label{sec:datasets}



\textbf{Synthetic Pattern Detection Tasks.} \hspace{0.2cm} 
The AML subgraph patterns seen in \cref{fig:aml_patterns} are used to create a controllable testbed of synthetic pattern detection tasks.
The key design principle is to ensure that the desired subgraph patterns appear randomly, rather than being inserted post hoc into a graph. The problem with inserting patterns is that it skews the random distribution, and simple indicators (such as the degrees of nodes) can be enough to solve the task approximately. 
For example, consider the extreme case of generating a random $k$-regular graph and then inserting a pattern. 
Nodes belonging to the pattern could be identified by checking whether their degree exceeds $k$. 
Additionally, if only inserted patterns are labeled, then randomly occurring patterns will be overlooked. 

To ensure that the desired subgraph patterns appear randomly, we introduce the \emph{random circulant graph} generator. Details of the generator and pseudocode can be found 
\iftoggle{arxiv}{%
    in \cref{app:synth_graph_generator}.
}{%
    in the appendix.
} 
The pattern detection tasks include degree-in/out (number of in/out edges), fan-in/out (number of unique in/out neighbors), scatter-gather, directed biclique, and directed cycles of length up to six.
Detailed descriptions can be found 
\iftoggle{arxiv}{%
    in \cref{app:synth_tasks}.
}{%
    in the appendix.
} 



\noindent
\textbf{Anti-Money Laundering (AML).} \hspace{0.2cm} 
Given the strict privacy regulations around financial data, real-world datasets are not readily available.
Instead, we use simulated money laundering data \citep{altman2023realistic}. The simulator behind these datasets generates a financial transaction network by modeling agents 
(banks, companies, and individuals) 
in a virtual world. 
The generator uses well-established laundering patterns to add realistic money laundering (illicit) transactions. 
We use two small and two medium-sized datasets, one of each with a higher illicit ratio (HI) and with a lower illicit ratio (LI). 
The dataset sizes and illicit ratios are provided 
\iftoggle{arxiv}{%
    in \cref{table:aml_eth_data} in the appendix. 
}{%
    in the appendix.
}
We use a 60-20-20 temporal train-validation-test split, i.e., we split the transactions after ordering them by their timestamps. 
Details can be found 
\iftoggle{arxiv}{%
    in \cref{app:aml_eth_data}.
}{%
    in the appendix.
}


\noindent
\textbf{Ethereum Phishing Detection (ETH).} \hspace{0.2cm} 
Since banks do not release their data, we turn to cryptocurrencies for a real-world dataset.
We use an Ethereum transaction network published on Kaggle \citep{xblockEthereum}, where some nodes are labeled as phishing accounts. 
We use a temporal train-validation-test split, but this time splitting the nodes. 
We use a 65-15-20 split because the illicit accounts are skewed towards the end of the dataset.
More details and dataset statistics can be found 
\iftoggle{arxiv}{%
    in \cref{app:aml_eth_data}.
}{%
    in the appendix.
}

\noindent
\textbf{Real-World Directed Graph Datasets.} \hspace{0.2cm} 
The theory results and the subgraph detection tasks demonstrate the general purpose potential of the architectural adaptations. 
However, testing our model on real-world benchmark datasets is important to further support these claims. 
For lack of established directed multi-graph benchmarks, we have taken three directed graph datasets, Chameleon, Squirrel \citep{pei2020geomPlusSquirrel}, and Arxiv-Year \citep{hu2020openArxiv}, and compare our approach with the state-of-the-art model for these benchmarks \citep{rusch2022gradient}. As these datasets are not the focus of this paper, we leave the experimental details and results to the appendix; 
\iftoggle{arxiv}{%
    please see \cref{app:real_world_benchmarks}.
}{%
    please see Appendix G.
}

\section{Experimental Setup}
\label{sec:setup}

\textbf{Base GNNs and Baselines.} \hspace{0.2cm} GIN with edge features \citep{hu2019strategies} is used as the main GNN base model with our adaptations added on top. GAT \citep{velivckovic2017graph} and PNA \citep{velickovic2019deep} are also used as base models, and we refer to their adapted versions as Multi-GAT and Multi-PNA, respectively. All three are also considered baselines. 
Additionally, GIN with ego IDs can be considered an ID-GNN \citep{you2021identity} baseline, and GIN with port numbering can be considered a CPNGNN \citep{sato2019approximation_port_numbers} baseline.
Since AML is an edge classification problem, we also include a baseline using edge updates \citep{battaglia2018relational_edgeupdates}, denoted GIN+EU. This approach is similar to replacing edges with nodes and running a GNN on said \emph{line graph}, which recently achieved state-of-the-art (SOTA) results in self-supervised money laundering detection \citep{cardoso2022laundrograph}.
We also include R-GCN \citep{schlichtkrull2018modelingRelGNN} as a baseline.
We do not focus on including a more expansive range of GNN baselines, for the simple reason that without (the proposed) adaptations, they are not equipped to deal with directed multigraphs. 
However, some additional results with ``more expressive'' GNNs can be found 
\iftoggle{arxiv}{%
    in \cref{app:add_gnn_baselines}.
}{%
    in the appendix.
}
As far as we are aware, there are no other GNNs that one could expect to achieve SOTA results on directed multigraphs.

We include a baseline representing the parallel line of work in financial crime detection that uses pre-calculated graph-based features (GFs) and tree-based classifiers to classify nodes or edges individually. 
We train XGBoost \citep{chen2016xgboost} and LightGBM \citep{ke2017lightgbm} models on the individual edges (or nodes) using the original raw features combined with additional graph-based features. 
This approach has produced SOTA results in financial applications \citep{weber2019anti, lo2022inspection_bitcoin}.

Given the size of the AML and ETH datasets, we use neighborhood sampling \citep{hamilton2017inductive} for all GNN-based models. 
Further details 
of the experimental setup for the different datasets 
can be found 
\iftoggle{arxiv}{%
    in \cref{app:exp_setup_details}.
}{%
    in the appendix.
}

\textbf{Scoring.} \hspace{0.2cm} Since we have very imbalanced datasets, accuracy and other popular metrics are not suitable. Instead, we use the minority class F1 score. This aligns well with what banks and regulators use in real-world scenarios.

\section{Results}
\label{sec:results}

\subsection{Synthetic Pattern Detection Results}
\label{subsec:subgraph_pattern_results}

The synthetic pattern detection results can be seen in \cref{tab:sim_results}.
The degree-out results reveal that the standard message-passing GNNs are unable to solve the degree-out task, achieving F1 scores below $44\%$. However, all the GNNs that are equipped with reverse MP score above $98\%$, thus supporting \cref{thm:reverseMP_outdegree}. 
The next column shows that port numbering is the critical adaptation for solving fan-in, though the F1 score is quite high even for the baseline GIN. On the other hand, for the fan-out task, the combination of reverse MP and port numbering is needed to score above $99\%$. Again, these results support \cref{thm:ports_fanin,thm:fanout}.
The ablation study of cumulative adaptations on top of GIN also supports \cref{cor:universality}: The combination of reverse MP, port numbering, and ego IDs, scores high on all of the subtasks, with only 6-cycle detection coming in below $90\%$. We see similar results when using other base GNN models, with Multi-PNA achieving the best overall results. Moreover, on the more complex tasks --- directed cycle, scatter-gather, and biclique detection --- the combination of the three is what leads to the first significant improvement in F1 scores. 
In the most extreme case, scatter-gather detection, the minority class F1 score jumps from $67.84\%$ with only reverse MP and port numbers to $97.42\%$ when ego IDs are added. No adaptation alone comes close to this score, so it is clear that the combination is needed. Similar jumps can be seen for directed 4-, 5-, and 6-cycle, and biclique detection. 
Increasing the dataset size and restricting the task to only the ``complex'' subtasks further increases the scores, with 6-cycle detection also reaching above $97\%$. 
\iftoggle{arxiv}{%
    More details can be found in \cref{app:complex_tasks}. Additional ablations can also be found in the appendix.
}{%
    More details can be found in the appendix, along with additional ablations.
}
In particular, we rerun the experiments using random unique node IDs as input features and see that node IDs are unable to replace port numbers and ego IDs in practice.


\begin{table*}
    \centering
    \begingroup
    \resizebox{\linewidth}{!}{%
    \begin{tabular}{lccccccccccc}
    \toprule
    Model & deg-in & deg-out & fan-in & fan-out & C2 & C3 & C4 & C5 & C6 & S-G & B-C \\
    \midrule
    GIN \citep{xu2018powerful, hu2019strategies} & 
    \cellcolor[HTML]{00451c} \color[HTML]{f1f1f1} 99.77 & \cellcolor[HTML]{f7fcf5} \color[HTML]{000000} 43.58 & \cellcolor[HTML]{00682a} \color[HTML]{f1f1f1} 95.57 & \cellcolor[HTML]{f7fcf5} \color[HTML]{000000} 35.91 & \cellcolor[HTML]{f7fcf5} \color[HTML]{000000} 34.67 & \cellcolor[HTML]{f7fcf5} \color[HTML]{000000} 58.00 & \cellcolor[HTML]{f7fcf5} \color[HTML]{000000} 50.80 & \cellcolor[HTML]{f7fcf5} \color[HTML]{000000} 43.12 & \cellcolor[HTML]{f7fcf5} \color[HTML]{000000} 48.59 & \cellcolor[HTML]{cbebc5} \color[HTML]{000000} 69.31 & \cellcolor[HTML]{ecf8e8} \color[HTML]{000000} 63.12 \\
    GAT \citep{velivckovic2017graph} & 
    \cellcolor[HTML]{f7fcf5} \color[HTML]{000000} 10.33 & \cellcolor[HTML]{f7fcf5} \color[HTML]{000000} 10.53 & \cellcolor[HTML]{f7fcf5} \color[HTML]{000000}  9.69 & \cellcolor[HTML]{f7fcf5} \color[HTML]{000000}  0.00 & \cellcolor[HTML]{f7fcf5} \color[HTML]{000000}  0.00 & \cellcolor[HTML]{f7fcf5} \color[HTML]{000000}  0.00 & \cellcolor[HTML]{f7fcf5} \color[HTML]{000000} 25.86 & \cellcolor[HTML]{f7fcf5} \color[HTML]{000000}  0.00 & \cellcolor[HTML]{f7fcf5} \color[HTML]{000000}  0.00 & \cellcolor[HTML]{f7fcf5} \color[HTML]{000000}  0.00 & \cellcolor[HTML]{f7fcf5} \color[HTML]{000000}  0.00 \\
    PNA \citep{velickovic2019deep} & 
    \cellcolor[HTML]{00471c} \color[HTML]{f1f1f1} 99.63 & \cellcolor[HTML]{f7fcf5} \color[HTML]{000000} 43.02 & \cellcolor[HTML]{006c2c} \color[HTML]{f1f1f1} 95.00 & \cellcolor[HTML]{f7fcf5} \color[HTML]{000000} 38.93 & \cellcolor[HTML]{f7fcf5} \color[HTML]{000000} 25.77 & \cellcolor[HTML]{f7fcf5} \color[HTML]{000000} 54.75 & \cellcolor[HTML]{f7fcf5} \color[HTML]{000000} 51.92 & \cellcolor[HTML]{f7fcf5} \color[HTML]{000000} 48.79 & \cellcolor[HTML]{f7fcf5} \color[HTML]{000000} 48.40 & \cellcolor[HTML]{e0f3db} \color[HTML]{000000} 65.88 & \cellcolor[HTML]{e2f4dd} \color[HTML]{000000} 65.51 \\
    GIN+EU \citep{battaglia2018relational_edgeupdates} &
    \cellcolor[HTML]{00491d} \color[HTML]{f1f1f1} 99.30 & \cellcolor[HTML]{f7fcf5} \color[HTML]{000000} 42.74 & \cellcolor[HTML]{006729} \color[HTML]{f1f1f1} 95.70 & \cellcolor[HTML]{f7fcf5} \color[HTML]{000000} 39.13 & \cellcolor[HTML]{f7fcf5} \color[HTML]{000000} 32.58 & \cellcolor[HTML]{f7fcf5} \color[HTML]{000000} 55.91 & \cellcolor[HTML]{f7fcf5} \color[HTML]{000000} 54.65 & \cellcolor[HTML]{f7fcf5} \color[HTML]{000000} 47.62 & \cellcolor[HTML]{f7fcf5} \color[HTML]{000000} 49.68 & \cellcolor[HTML]{d0edca} \color[HTML]{000000} 68.54 & \cellcolor[HTML]{e7f6e2} \color[HTML]{000000} 64.64 \\
    \midrule
    GIN+EgoIDs \citep{you2021identity} & 
    \cellcolor[HTML]{00451c} \color[HTML]{f1f1f1} 99.78 & \cellcolor[HTML]{f7fcf5} \color[HTML]{000000} 51.48 & \cellcolor[HTML]{006c2c} \color[HTML]{f1f1f1} 95.06 & \cellcolor[HTML]{f7fcf5} \color[HTML]{000000} 49.24 & \cellcolor[HTML]{005221} \color[HTML]{f1f1f1} 98.13 & \cellcolor[HTML]{005321} \color[HTML]{f1f1f1} 97.97 & \cellcolor[HTML]{f7fcf5} \color[HTML]{000000} 53.12 & \cellcolor[HTML]{f7fcf5} \color[HTML]{000000} 44.37 & \cellcolor[HTML]{f7fcf5} \color[HTML]{000000} 45.42 & \cellcolor[HTML]{dcf2d7} \color[HTML]{000000} 66.44 & \cellcolor[HTML]{e9f7e5} \color[HTML]{000000} 63.90 \\
    GIN+Ports \citep{sato2019approximation_port_numbers} & 
    \cellcolor[HTML]{00481d} \color[HTML]{f1f1f1} 99.47 & \cellcolor[HTML]{f7fcf5} \color[HTML]{000000} 45.00 & \cellcolor[HTML]{00471c} \color[HTML]{f1f1f1} 99.59 & \cellcolor[HTML]{f7fcf5} \color[HTML]{000000} 41.51 & \cellcolor[HTML]{f7fcf5} \color[HTML]{000000} 27.79 & \cellcolor[HTML]{f7fcf5} \color[HTML]{000000} 56.11 & \cellcolor[HTML]{f7fcf5} \color[HTML]{000000} 42.68 & \cellcolor[HTML]{f7fcf5} \color[HTML]{000000} 41.11 & \cellcolor[HTML]{f7fcf5} \color[HTML]{000000} 44.99 & \cellcolor[HTML]{d3eecd} \color[HTML]{000000} 67.99 & \cellcolor[HTML]{e1f3dc} \color[HTML]{000000} 65.76 \\
    \midrule
    GIN+ReverseMP \citep{jaume2019edgnnbidirecibmmaria} & 
    \cellcolor[HTML]{004d1f} \color[HTML]{f1f1f1} 98.87 & \cellcolor[HTML]{004a1e} \color[HTML]{f1f1f1} 99.08 & \cellcolor[HTML]{006d2c} \color[HTML]{f1f1f1} 94.99 & \cellcolor[HTML]{006b2b} \color[HTML]{f1f1f1} 95.25 & \cellcolor[HTML]{f7fcf5} \color[HTML]{000000} 35.96 & \cellcolor[HTML]{e9f7e5} \color[HTML]{000000} 63.85 & \cellcolor[HTML]{ccebc6} \color[HTML]{000000} 69.09 & \cellcolor[HTML]{d7efd1} \color[HTML]{000000} 67.44 & \cellcolor[HTML]{bee5b8} \color[HTML]{000000} 71.23 & \cellcolor[HTML]{e0f3db} \color[HTML]{000000} 65.83 & \cellcolor[HTML]{def2d9} \color[HTML]{000000} 66.18 \\
    {\ +Ports} & 
    \cellcolor[HTML]{005120} \color[HTML]{f1f1f1} 98.41 & \cellcolor[HTML]{005120} \color[HTML]{f1f1f1} 98.35 & \cellcolor[HTML]{005020} \color[HTML]{f1f1f1} 98.51 & \cellcolor[HTML]{004a1e} \color[HTML]{f1f1f1} 99.16 & \cellcolor[HTML]{f7fcf5} \color[HTML]{000000} 39.15 & \cellcolor[HTML]{ebf7e7} \color[HTML]{000000} 63.58 & \cellcolor[HTML]{cdecc7} \color[HTML]{000000} 69.00 & \cellcolor[HTML]{c4e8bd} \color[HTML]{000000} 70.35 & \cellcolor[HTML]{a0d99b} \color[HTML]{000000} 75.04 & \cellcolor[HTML]{d4eece} \color[HTML]{000000} 67.84 & \cellcolor[HTML]{e1f3dc} \color[HTML]{000000} 65.78 \\
    {\ +EgoIDs (Multi-GIN)} & 
    \cellcolor[HTML]{00481d} \color[HTML]{f1f1f1} 99.48 & \cellcolor[HTML]{004a1e} \color[HTML]{f1f1f1} 99.09 & \cellcolor[HTML]{00471c} \color[HTML]{f1f1f1} 99.62 & \cellcolor[HTML]{00491d} \color[HTML]{f1f1f1} 99.32 & \cellcolor[HTML]{004c1e} \color[HTML]{f1f1f1} 98.97 & \cellcolor[HTML]{004e1f} \color[HTML]{f1f1f1} 98.73 & \cellcolor[HTML]{005924} \color[HTML]{f1f1f1} 97.46 & \cellcolor[HTML]{17813d} \color[HTML]{f1f1f1} 91.60 & \cellcolor[HTML]{48ae60} \color[HTML]{f1f1f1} 84.23 & \cellcolor[HTML]{005924} \color[HTML]{f1f1f1} 97.42 & \cellcolor[HTML]{05712f} \color[HTML]{f1f1f1} 94.33 \\
    \midrule
    Multi-GAT & 
    \cellcolor[HTML]{004e1f} \color[HTML]{f1f1f1} 98.68 & \cellcolor[HTML]{005120} \color[HTML]{f1f1f1} 98.36 & \cellcolor[HTML]{00491d} \color[HTML]{f1f1f1} 99.28 & \cellcolor[HTML]{00491d} \color[HTML]{f1f1f1} 99.33 & \cellcolor[HTML]{004e1f} \color[HTML]{f1f1f1} 98.61 & \cellcolor[HTML]{004c1e} \color[HTML]{f1f1f1} 98.93 & \cellcolor[HTML]{004d1f} \color[HTML]{f1f1f1} 98.90 & \cellcolor[HTML]{006529} \color[HTML]{f1f1f1} 95.82 & \cellcolor[HTML]{16803c} \color[HTML]{f1f1f1} 91.81 & \cellcolor[HTML]{005f26} \color[HTML]{f1f1f1} 96.66 & \cellcolor[HTML]{359e53} \color[HTML]{f1f1f1} 86.92 \\
    Multi-PNA & 
    \cellcolor[HTML]{00471c} \color[HTML]{f1f1f1} 99.64 & \cellcolor[HTML]{00491d} \color[HTML]{f1f1f1} 99.25 & \cellcolor[HTML]{00481d} \color[HTML]{f1f1f1} 99.53 & \cellcolor[HTML]{00481d} \color[HTML]{f1f1f1} 99.41 & \cellcolor[HTML]{00451c} \color[HTML]{f1f1f1} 99.71 & \cellcolor[HTML]{00471c} \color[HTML]{f1f1f1} 99.54 & \cellcolor[HTML]{00481d} \color[HTML]{f1f1f1} 99.49 & \cellcolor[HTML]{005924} \color[HTML]{f1f1f1} 97.46 & \cellcolor[HTML]{2a924a} \color[HTML]{f1f1f1} 88.75 & \cellcolor[HTML]{004a1e} \color[HTML]{f1f1f1} 99.07 & \cellcolor[HTML]{005e26} \color[HTML]{f1f1f1} 96.77 \\
    Multi-GIN+EU & 
    \cellcolor[HTML]{00471c} \color[HTML]{f1f1f1} 99.55 & \cellcolor[HTML]{00481d} \color[HTML]{f1f1f1} 99.53 & \cellcolor[HTML]{00451c} \color[HTML]{f1f1f1} 99.76 & \cellcolor[HTML]{00451c} \color[HTML]{f1f1f1} 99.77 & \cellcolor[HTML]{00491d} \color[HTML]{f1f1f1} 99.37 & \cellcolor[HTML]{00451c} \color[HTML]{f1f1f1} 99.71 & \cellcolor[HTML]{004e1f} \color[HTML]{f1f1f1} 98.73 & \cellcolor[HTML]{006729} \color[HTML]{f1f1f1} 95.73 & \cellcolor[HTML]{2d954d} \color[HTML]{f1f1f1} 88.38 & \cellcolor[HTML]{004d1f} \color[HTML]{f1f1f1} 98.81 & \cellcolor[HTML]{005522} \color[HTML]{f1f1f1} 97.82 \\
    \bottomrule
    \end{tabular}
    }
    \endgroup
    \caption{Minority class F1 scores ($\%$) for the synthetic subgraph detection tasks. First from the top are the standard MPNN baselines; then the results with each adaptation added separately on top of GIN; followed by GIN with the adaptations added cumulatively; and finally, results for the other GNN baselines with the three adaptations (Multi-GNNs).  The C$k$ abbreviations stand for directed $k$-cycle detection, S-G stands for scatter-gather and B-C stands for biclique detection. We report minority class F1 scores averaged over five runs. We omit standard deviations in favor of readability.
    }
    \label{tab:sim_results}
\end{table*}

\subsection{AML Results}

\begin{table*}[t!]
    \centering
    \begingroup
    \resizebox{\linewidth}{!}{%
    \begin{tabular}{lccccc}
    \toprule
    {Model} & 
    AML Small HI & 
    AML Small LI & 
    AML Medium HI & 
    AML Medium LI & 
    ETH
    \\
    \midrule
    
    LightGBM+GFs \citep{altman2023realistic} & 
    \cellcolor[HTML]{0a7633} \color[HTML]{f1f1f1} $62.86 \pm 0.25$ & 
    \cellcolor[HTML]{bee5b8} \color[HTML]{000000} $20.83 \pm 1.50$ & 
    \cellcolor[HTML]{16803c} \color[HTML]{f1f1f1} $59.48 \pm 0.15$ & 
    \cellcolor[HTML]{bee5b8} \color[HTML]{000000} $20.85 \pm 0.38$ &
    \cellcolor[HTML]{2d954d} \color[HTML]{f1f1f1} $53.20 \pm 0.60$ 
    \\
    XGBoost+GFs \citep{altman2023realistic} & 
    \cellcolor[HTML]{097532} \color[HTML]{f1f1f1} $63.23 \pm 0.17$ & 
    \cellcolor[HTML]{a4da9e} \color[HTML]{000000} $27.30 \pm 0.33$ & 
    \cellcolor[HTML]{006c2c} \color[HTML]{f1f1f1} $65.70 \pm 0.26$ & 
    \cellcolor[HTML]{a0d99b} \color[HTML]{000000} $28.16 \pm 0.14$ &
    \cellcolor[HTML]{39a257} \color[HTML]{f1f1f1} $49.40 \pm 0.54$
    \\

    GIN \citep{xu2018powerful, hu2019strategies} & 
    \cellcolor[HTML]{9fd899} \color[HTML]{000000} $28.70 \pm 1.13$ & 
    \cellcolor[HTML]{e8f6e4} \color[HTML]{000000} $7.90 \pm 2.78$ & 
    \cellcolor[HTML]{5ab769} \color[HTML]{f1f1f1} $42.20 \pm 0.44$ & 
    \cellcolor[HTML]{f0f9ec} \color[HTML]{000000} $3.86 \pm 3.62$ & 
    \cellcolor[HTML]{a7dba0} \color[HTML]{000000} $26.92 \pm 7.52$
    \\
    PNA \citep{velickovic2019deep} & 
    \cellcolor[HTML]{218944} \color[HTML]{f1f1f1} $56.77 \pm 2.41$ & 
    \cellcolor[HTML]{d4eece} \color[HTML]{000000} $14.85 \pm 1.46$ & 
    \cellcolor[HTML]{16803c} \color[HTML]{f1f1f1} $59.71 \pm 1.91$ & 
    \cellcolor[HTML]{a3da9d} \color[HTML]{000000} $27.73 \pm 1.65$ & 
    \cellcolor[HTML]{329b51} \color[HTML]{f1f1f1} $51.49 \pm 4.26$
    \\
    GIN+EU \citep{battaglia2018relational_edgeupdates} & 
    \cellcolor[HTML]{3fa85b} \color[HTML]{f1f1f1} $47.73 \pm 7.86$ & 
    \cellcolor[HTML]{c0e6b9} \color[HTML]{000000} $20.62 \pm 2.41$ & 
    \cellcolor[HTML]{39a257} \color[HTML]{f1f1f1} $49.26 \pm 4.02$ & 
    \cellcolor[HTML]{ebf7e7} \color[HTML]{000000} $6.19 \pm 8.32$ & 
    \cellcolor[HTML]{86cc85} \color[HTML]{000000} $33.92 \pm 7.34$
    \\
    R-GCN \citep{schlichtkrull2018modelingRelGNN} &
    \cellcolor[HTML]{5db96b} \color[HTML]{f1f1f1} $41.78 \pm 0.48$ & 
    \cellcolor[HTML]{e9f7e5} \color[HTML]{000000} $7.43 \pm 0.38$ & 
    OOM & 
    OOM & 
    OOM 
    \\

    \midrule
    
    GIN+EgoIDs \citep{you2021identity} & 
    \cellcolor[HTML]{68be70} \color[HTML]{000000} $39.65 \pm 4.73$ & 
    \cellcolor[HTML]{d3eecd} \color[HTML]{000000}  $14.98 \pm 2.66$ & 
    \cellcolor[HTML]{4aaf61} \color[HTML]{f1f1f1} $45.26 \pm 2.16$ & 
    \cellcolor[HTML]{dff3da} \color[HTML]{000000} $11.17 \pm 6.41$ & 
    \cellcolor[HTML]{aadda4} \color[HTML]{000000} $26.01 \pm 2.27$
    \\
    GIN+Ports \citep{sato2019approximation_port_numbers} & 
    \cellcolor[HTML]{278f48} \color[HTML]{f1f1f1} $54.85 \pm 0.89$ & 
    \cellcolor[HTML]{bce4b5} \color[HTML]{000000} $21.41 \pm 2.40$ & 
    \cellcolor[HTML]{29914a} \color[HTML]{f1f1f1} $54.22 \pm 1.94$ & 
    \cellcolor[HTML]{e2f4dd} \color[HTML]{000000} $10.51 \pm 12.82$& 
    \cellcolor[HTML]{8ace88} \color[HTML]{000000} $32.96 \pm 0.25$
    \\

    \midrule
    
    GIN+ReverseMP \citep{jaume2019edgnnbidirecibmmaria} & 
    \cellcolor[HTML]{42ab5d} \color[HTML]{f1f1f1} $46.79 \pm 4.97$ & 
    \cellcolor[HTML]{d0edca} \color[HTML]{000000} $15.98 \pm 4.39$ & 
    \cellcolor[HTML]{309950} \color[HTML]{f1f1f1} $51.93 \pm 2.90$ & 
    \cellcolor[HTML]{d7efd1} \color[HTML]{000000} $14.00 \pm 9.34$ & 
    \cellcolor[HTML]{78c679} \color[HTML]{000000} $36.86 \pm 8.12$
    \\
    {\ +Ports} & 
    \cellcolor[HTML]{208843} \color[HTML]{f1f1f1} $56.85 \pm 2.64$ & 
    \cellcolor[HTML]{b2e0ac} \color[HTML]{000000} $23.80 \pm 4.07$ & 
    \cellcolor[HTML]{1f8742} \color[HTML]{f1f1f1} $57.15 \pm 0.76$ & 
    \cellcolor[HTML]{dff3da} \color[HTML]{000000} $11.39 \pm 8.36$ & 
    \cellcolor[HTML]{58b668} \color[HTML]{f1f1f1} $42.51 \pm 7.16$
    \\
    {\ +EgoIDs (Multi-GIN)} & 
    \cellcolor[HTML]{1f8742} \color[HTML]{f1f1f1} $57.15 \pm 4.99$ & 
    \cellcolor[HTML]{bae3b3} \color[HTML]{000000} $22.12 \pm 2.88$ & 
    \cellcolor[HTML]{238b45} \color[HTML]{f1f1f1} $56.23 \pm 1.51$ & 
    \cellcolor[HTML]{d5efcf} \color[HTML]{000000} $14.55 \pm 2.91$ & 
    \cellcolor[HTML]{56b567} \color[HTML]{f1f1f1} $42.86 \pm 2.53$
    \\

    \midrule

    Multi-GIN+EU & 
    \cellcolor[HTML]{026f2e} \color[HTML]{f1f1f1} $64.79 \pm 1.22$ & 
    \cellcolor[HTML]{a7dba0} \color[HTML]{000000} $26.88 \pm 6.63$ & 
    \cellcolor[HTML]{18823d} \color[HTML]{f1f1f1} $58.92 \pm 1.83$ & 
    \cellcolor[HTML]{cfecc9} \color[HTML]{000000} $16.30 \pm 4.73$ & 
    \cellcolor[HTML]{3ca559} \color[HTML]{f1f1f1} $48.37 \pm 6.62$
    \\
    Multi-PNA & 
    \cellcolor[HTML]{006c2c} \color[HTML]{f1f1f1} $64.59 \pm 3.60$ & 
    \cellcolor[HTML]{98d594} \color[HTML]{000000} $30.65 \pm 2.00$ & 
    \cellcolor[HTML]{006c2c} \color[HTML]{f1f1f1} $65.67 \pm 2.66$ & 
    \cellcolor[HTML]{88ce87} \color[HTML]{000000} $33.23 \pm 1.31$ & 
    \cellcolor[HTML]{016e2d} \color[HTML]{f1f1f1} $65.28 \pm 2.89$
    \\
    Multi-PNA+EU & 
    \cellcolor[HTML]{006227} \color[HTML]{f1f1f1} $68.16 \pm 2.65$ & 
    \cellcolor[HTML]{8ace88} \color[HTML]{000000} $33.07 \pm 2.63$ & 
    \cellcolor[HTML]{00692a} \color[HTML]{f1f1f1} $66.48 \pm 1.63$ & 
    \cellcolor[HTML]{7ac77b} \color[HTML]{000000} $36.07 \pm 1.17$ & 
    \cellcolor[HTML]{00682a} \color[HTML]{f1f1f1} $66.58 \pm 1.60$
    \\
    
    \bottomrule
    \end{tabular}
    }
    \endgroup
    \caption{
    Minority class F1 scores ($\%$) for the AML and ETH tasks. 
    HI indicates a higher illicit ratio and LI indicates a lower illicit ratio.
    The models are organized as in \cref{tab:sim_results}. ``OOM'' indicates that the model ran out of GPU memory.}
    \label{tab:full_aml_eth_results}
\end{table*}

The results for the AML datasets can be seen in \cref{tab:full_aml_eth_results}. 
For AML Small HI, we see that our adaptations boost the minority class F1 score of GIN from $28.7\%$ to $57.2\%$, a gain of almost $30\%$. The largest improvements are brought by reverse MP and port numbering, taking the F1 score from $28.7\%$ to $56.9\%$, whilst ego IDs do not make much difference here. 
The results for the other AML datasets show a similar trend with overall gains of $14.2\%$, $14.0\%$, and $10.7\%$ for GIN, again with diminishing returns as more adaptations are added. 
The two rows corresponding to port numbering --- \textit{GIN+Ports} and \textit{+Ports} --- indicate clear gains from using port numbering, both when used alone and on top of reverse MP. 
The support for ego IDs is less clear, with clear gains when used as an individual adaptation but no significant gains when added on top of reverse MP and port numbering.
\iftoggle{arxiv}{%
    Note that only two GNN layers were used, so this conclusion could change as more layers are added and longer cycles can also be detected. 
}

The full set of adaptations was tested with three other base models, GIN+EU (GIN with edge updates), PNA, and PNA+EU. In each case, and across almost all AML datasets, we see clear gains from using the adaptations. These gains underline the effectiveness and versatility of the approach. 
We further note that Multi-PNA+EU outperforms all of the baselines on all of the AML datasets.
This is particularly impressive when compared with the tree-based methods using graph-based features (XGBoost+GFs and LightGBM+GFs) since the hand-crafted features align perfectly with the illicit money laundering patterns used by the simulator. Moreover, these tree-based methods have been SOTA in previous financial applications \citep{weber2019anti, lo2022inspection_bitcoin}.


Recall scores for individual money laundering patterns can be found 
\iftoggle{arxiv}{%
    in \cref{app:detailed_aml_results}.
}{%
    in the appendix.
} 
It is worth noting that the majority of the illicit transactions that belong to money laundering patterns are identified, and the overall dataset scores are greatly influenced by the proportion of lone (not belonging to a money laundering pattern) illicit transactions in the datasets. Lone illicit transactions are very difficult to identify. 

For training times and inference throughput rates of models based on GIN, 
\iftoggle{arxiv}{%
    please see \cref{tab:runtimes} in the appendix.
}{%
    please see the appendix.
}
Notably, with all the adaptations, the inference rate of Multi-GIN still surpasses 18k transactions per second on a single GPU.



\subsection{ETH Results}

Finally, we test our adaptations on a real-world financial crime dataset --- Ethereum phishing account classification. The results are provided in \cref{tab:full_aml_eth_results}. 
Similar to the AML datasets, we see a consistent improvement in final scores as we add the adaptations. In total, the minority class F1 score jumps from $26.9\%$ without adaptations to $42.9\%$ with reverse MP, port numbering, and ego IDs. Again, the largest single improvement is due to the reverse MP. 
In this case, Multi-GIN does not outperform all of the baselines, but the adaptations also significantly boost PNA performance, and Multi-PNA and Multi-PNA+EU beat all the baselines by more than $12\%$.







\section{Conclusion}
\label{sec:conclusion}


This work has investigated a series of straightforward adaptations capable of transforming conventional message-passing GNNs into provably powerful directed multigraph learners. Our contributions to the field of graph neural networks are threefold.
Firstly, our theoretical analysis addresses a notable gap in the existing literature about the power of combining different GNN adaptations/augmentations. Specifically, we prove that ego IDs combined with port numbering and reverse message passing enable a suitably powerful message-passing GNN, such as GIN, to compute unique node IDs and therefore detect any directed subgraph patterns. 
Secondly, our theoretical findings are validated empirically with a range of synthetic subgraph detection tasks. The practical results closely mirror the theoretical expectations, confirming that the combination of all three adaptations is needed to detect the more complex subgraphs.
Lastly, we show how our adaptations can be applied to two important financial crime problems: detecting money laundering transactions and phishing accounts. GNNs enhanced with our proposed adaptations achieve impressive results in both tasks, either matching or surpassing relevant baselines. Reverse message passing and port numbering again prove crucial in reaching the highest scores, however, we find that 
ego IDs do not provide much additional benefit for these datasets.

Although this work has focused on financial crime applications, the theory and practical results have a broader relevance. Immediate future work could involve exploring applications of our methods to other directed multigraph problems. An initial exploration can be found in the appendix, showing promising results on three real-world datasets. However, further experiments are needed to confirm general applicability in various domains. Additionally, future work could explore the relationship between the computational complexity of different subgraph detection problems and GNN performance.



\section*{Acknowledgements}

The support of the Swiss National Science Foundation (project numbers: 172610 and 212158) for this work is gratefully acknowledged.

{\small
\bibliography{aaai24}
}

\iftoggle{arxiv}{%
\clearpage

\appendix

\section{Motivating Examples}

\subsection{Example Financial Transaction Graph}
\label{app:intro_example}

\begin{figure}[ht]
\centering
\begin{tikzpicture}[scale=0.8, every node/.style={scale=1.0}, >=latex]
    \node[businessman, female, draw, minimum size=0.4cm] (P1) at  (-6,1) {};
    \node[businessman, female, draw, minimum size=0.4cm] (P2) at  (-6,-1) {};
    \node[businessman, female, evil, draw, minimum size=0.4cm] (P3) at  (-3,-2) {};
    \node[circle, draw, minimum size=0.4cm] (A) at  (-3,0) {};
    \node[circle, draw, minimum size=0.4cm] (B1) at  (0,1.5) {};
    \node[circle, draw, minimum size=0.4cm] (B2) at  (0,0) {};
    \node[circle, draw, minimum size=0.4cm] (B3) at  (0,-1.5) {};
    \node[devil, draw, minimum size=0.4cm] (C) at  (3,0) {};
    \draw [semithick,->] (P1) edge node[pos=0.45,above] {\textcolor{blue}{\$}} (A);
    \draw [semithick,->] (P2) edge node[pos=0.45,above] {\textcolor{blue}{\$\$}} (A);
    \draw [semithick,->,dotted] (P3) edge node[pos=0.45] {\textcolor{red}{\$\$\$}} (A);
    \draw [line width=0.5mm,->,red] (A) edge node[pos=0.55,above] {\textcolor{blue}{\$}\textcolor{red}{\$}} (B1);
    \draw [line width=0.5mm,->,red] (A) edge node[pos=0.55,above] {\textcolor{blue}{\$}\textcolor{red}{\$}} (B2);
    \draw [line width=0.5mm,->,red] (A) edge node[pos=0.55,below] {\textcolor{blue}{\$}\textcolor{red}{\$}} (B3);
    \draw [line width=0.5mm,->,red] (B1) edge node[pos=0.45,above] {\textcolor{blue}{\$}\textcolor{red}{\$}} (C);
    \draw [line width=0.5mm,->,red] (B2) edge node[pos=0.45,above] {\textcolor{blue}{\$}\textcolor{red}{\$}} (C);
    \draw [line width=0.5mm,->,red] (B3) edge node[pos=0.45,below] {\textcolor{blue}{\$}\textcolor{red}{\$}} (C);
    \node at (A) [yshift=0.2cm,anchor=south,text=blue] {\Bed};
    \node at (P1) [yshift=-0.3cm,anchor=north] {Alice};
    \node at (P2) [yshift=-0.3cm,anchor=north] {Bob};
    \node at (P3) [yshift=-0.3cm,anchor=north] {Sean};
    \node at (C) [yshift=-0.3cm,anchor=north] {Tim};
\end{tikzpicture}
\caption{Example of money laundering in a network of financial transactions. Alice and Bob stay at a hotel, which is run by a criminal group headed by Tim. Sean has some dirty money (red dollars) from criminal activities that he wants to transfer to Tim. They use the hotel for laundering the money. They mix the dirty cash with clean money from guests, pay different contractors for supplies, and then transfer these payments to Tim. The money transfer from Sean to the hotel is a cash payment hidden from banks and financial authorities (dotted edge). However, the scatter-gather pattern (bold, red edges) could be revealing of a money laundering scheme.}
\label{fig:intro_ex}
\end{figure}
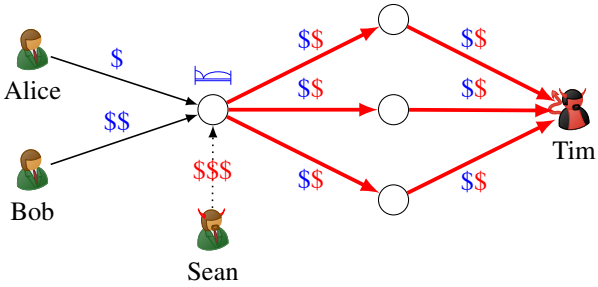

\subsection{Passing Port Numbers in Both Directions}
\label{app:bidirec_ports}

Note that standard port numbering as introduced by \citet{sato2019approximation_port_numbers} is not sufficient for assigning unique node IDs. 
Unlike \citet{sato2019approximation_port_numbers} attach only the local port numbers at a node to received messages. In contrast, we attach the port numbers in both directions, i.e., a node sees both the port number it has assigned to a neighbor and the port number that the neighbor has assigned to it. 
This turns out to be crucial for inferring unique node IDs.
For example, consider a simple undirected star graph with one internal node and 3 leaves, as shown in \cref{fig:ex_bidirec_ports}. The internal node assigns unique port numbers to the leaves, and according to \citep{sato2019approximation_port_numbers}, it concatenates each of these port numbers with the respective incoming messages from the three leaves, allowing it to distinguish the leaves and attribute the messages. However, the internal node’s port numbers are not attached to outgoing messages, and since the leaves have a single port (neighbor), they receive identical messages with the identical port number attached in each message passing round. So the leaves have no way of attaining unique IDs when the internal node is the centre node with the ego ID. Similarly, the two unmarked nodes have no way of attaining unique node IDs when the third leaf is the centre node. Our approach solves this problem by attaching the port numbers in both directions to the messages. 

\begin{figure}[t]
    \centering
    \begin{tikzpicture}[scale=0.5, every node/.style={scale=0.9}, >=latex]
    \node[circle, draw, minimum size=0.5cm, inner sep=0] (A) at  (0,3.6) {$a$};
    \node[circle, draw, minimum size=0.5cm, inner sep=0] (B) at  (0,1.8) {$b$};
    \node[circle, draw, minimum size=0.5cm, inner sep=0] (C) at  (0,0.0) {$c$};
    \node[circle, draw, minimum size=0.5cm, inner sep=0, fill=gray!30] (D) at  (4,1.8) {$d$};
    \draw [semithick] (A) -- (D);
    \draw [semithick] (B) -- (D);
    \draw [semithick] (C) -- (D);
    \node[blue] at (0.7,0) {\footnotesize{$1$}};
    \node[blue] at (0.7,1.8) {\footnotesize{$1$}};
    \node[blue] at (0.7,3.6) {\footnotesize{$1$}};
    \node[blue] at (3.25,2.5) {\footnotesize{$1$}};
    \node[blue] at (3.1,1.8) {\footnotesize{$2$}};
    \node[blue] at (3.25,1.1) {\footnotesize{$3$}};
    
    \node[circle, draw, minimum size=0.5cm, inner sep=0] (A2) at  (8,3.6) {$a'$};
    \node[circle, draw, minimum size=0.5cm, inner sep=0] (B2) at  (8,1.8) {$b'$};
    \node[circle, draw, minimum size=0.5cm, inner sep=0] (C2) at  (8,0) {$c'$};
    \node[circle, draw, minimum size=0.5cm, inner sep=0, fill=gray!30] (D2) at  (12,1.8) {$d'$};
    \draw [semithick] (A2) -- (D2) node[blue, pos=0.5,above] {\footnotesize{$(1,1)$}};
    \draw [semithick] (B2) -- (D2) node[blue, pos=0.3,above] {\footnotesize{$(1,2)$}};
    \draw [semithick] (C2) -- (D2) node[blue, pos=0.5,below] {\footnotesize{$(1,3)$}};
\end{tikzpicture}
    \caption{\citet{sato2019approximation_port_numbers} attach only the local port numbers to received messages, as indicated on the left. 
    Port numbers are indicated in blue.
    For example, node $c$ attaches port number $1$ to all messages from $d$.
    So in particular, nodes $a$, $b$, and $c$ can never be distinguished. We use the port numbers on either side on an edge as edge features, so both port numbers are seen by both incident nodes. In this example this might not be a problem --- $a$, $b$, and $c$ likely have the same ground truth labels --- but this means that unique node IDs cannot be generated/propagated.}
    \label{fig:ex_bidirec_ports}
\end{figure}
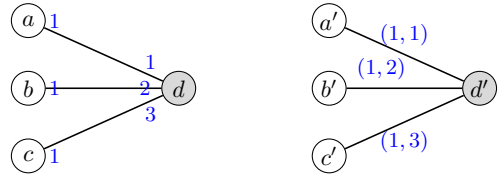

\subsection{Ego ID Cycle Detection Counterexample}

We give a counterexample to show that ego IDs alone \citep{you2021identity} are not sufficient to detect 6-cycles.
Consider nodes $u$ and $u'$ in \cref{fig:ego_fail_ex}. They would receive the same label when using a standard MPNN with ego IDs ($u$ and $u'$ being the center nodes), but $u'$ is part of several $5$-cycles and $u$ is not. To see intuitively why, note first that the blue edges form two $3$-cycles on the one side and a $6$-cycle on the other, and that the 1-WL test can not distinguish these alone. Now, when $u$ and $u'$ are the center nodes, the only extra information is that $u$ and $u'$ are marked, but since they are fully connected in their graphs, the mark does not help in distinguishing any of the other nodes, so the neighbors remain indistinguishable and indeed the two ego graphs remain indistinguishable. This means they could not both be labeled correctly in any task that relies on $5$-cycles.
Note that the predictions for any of the other respective nodes (e.g., $a$ and $a'$) will not be identical, since when these are the center nodes with the ego ID, then the GNN can detect that the ones on the left are in $3$-cycles, whereas the ones on the right are not.

\begin{figure}[t]
\centering
\begin{tikzpicture}[scale=0.35, every node/.style={scale=0.95}, >=latex]
    \node[circle, draw, minimum size=0.5cm, inner sep=0pt] (A) at  (2,3.5) {$a$};
    \node[circle, draw, minimum size=0.5cm, inner sep=0pt] (B) at  (4,0) {$b$};
    \node[circle, draw, minimum size=0.5cm, inner sep=0pt] (C) at  (2,-3.5) {$c$};
    \node[circle, draw, minimum size=0.5cm, inner sep=0pt] (D) at  (-2,-3.5) {$d$};
    \node[circle, draw, minimum size=0.5cm, inner sep=0pt] (E) at  (-4,0) {$e$};
    \node[circle, draw, minimum size=0.5cm, inner sep=0pt] (F) at  (-2,3.5) {$f$};
    \node[circle, draw, minimum size=0.5cm, inner sep=0pt, fill=gray!30] (U) at  (0,0) {$u$};
    \draw [semithick,->, blue] (A) -- (B);
    \draw [semithick,->, blue] (B) -- (C);
    \draw [semithick,->, blue] (C) -- (A);
    \draw [semithick,->, blue] (D) -- (E);
    \draw [semithick,->, blue] (E) -- (F);
    \draw [semithick,->, blue] (F) -- (D);
    \draw [semithick,<->, black] (A) -- (U);
    \draw [semithick,<->, black] (B) -- (U);
    \draw [semithick,<->, black] (C) -- (U);
    \draw [semithick,<->, black] (D) -- (U);
    \draw [semithick,<->, black] (E) -- (U);
    \draw [semithick,<->, black] (F) -- (U);
    
    \node[circle, draw, minimum size=0.5cm, inner sep=0pt] (A2) at  (14,3.5) {$a'$};
    \node[circle, draw, minimum size=0.5cm, inner sep=0pt] (B2) at  (16,0) {$b'$};
    \node[circle, draw, minimum size=0.5cm, inner sep=0pt] (C2) at  (14,-3.5) {$c'$};
    \node[circle, draw, minimum size=0.5cm, inner sep=0pt] (D2) at  (10,-3.5) {$d'$};
    \node[circle, draw, minimum size=0.5cm, inner sep=0pt] (E2) at  (8,0) {$e'$};
    \node[circle, draw, minimum size=0.5cm, inner sep=0pt] (F2) at  (10,3.5) {$f'$};
    \node[circle, draw, minimum size=0.5cm, inner sep=0pt, fill=gray!30] (U2) at  (12,0) {$u'$};
    \draw [semithick,->, blue] (A2) -- (B2);
    \draw [semithick,->, blue] (B2) -- (C2);
    \draw [semithick,->, blue] (C2) -- (D2);
    \draw [semithick,->, blue] (D2) -- (E2);
    \draw [semithick,->, blue] (E2) -- (F2);
    \draw [semithick,->, blue] (F2) -- (A2);
    \draw [semithick,<->, black] (A2) -- (U2);
    \draw [semithick,<->, black] (B2) -- (U2);
    \draw [semithick,<->, black] (C2) -- (U2);
    \draw [semithick,<->, black] (D2) -- (U2);
    \draw [semithick,<->, black] (E2) -- (U2);
    \draw [semithick,<->, black] (F2) -- (U2);
\end{tikzpicture}
\caption{Example nodes ($u$ and $u'$) with different $5$-cycle counts that are not distinguishable by a standard MPNN with ego IDs. For example, $(u', a', b', c', d', u')$ is a $5$-cycle on the right, but there are no $5$-cycles involving u on the left. The example also works with undirected edges.}
\label{fig:ego_fail_ex}
\end{figure}
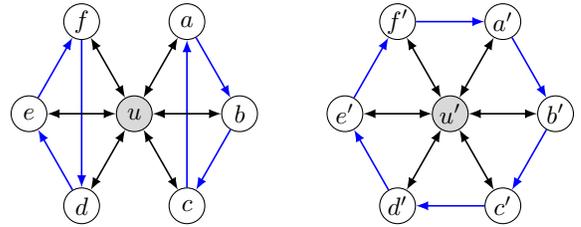

\section{Proofs}
\label{app:proofs}

\subsection{Proof of \cref{thm:reverseMP_outdegree}}

\begin{proof}
Consider a message-passing GNN with sum aggregation. Then with a single GNN layer, initial node features $h^{(0)}(v) = 1$ for all $v \in V(G)$, we have $a_{in}^{(t)}(v) = d_{in}(v)$ and $a_{out}^{(t)}(v) = d_{out}(v)$. Choosing $\textsc{Update} \left( h, a_{in}, a_{out} \right) = a_{out}$, we get $h^{(1)}(v) = a_{out}^{(t)}(v) = d_{out}(v)$.
\end{proof}

\subsection{Proof of \cref{thm:ports_fanin}}

\begin{proof}
Since consecutive integers are used to number the incoming ports, the max aggregation function can directly output the maximum of the incoming port numbers to give the desired result.  
\end{proof}

The statement also holds for MPNNs with sum aggregation. However, the proof is less intuitive and relies on the universality of sum aggregation for multisets (Lemma 5 from \citet{xu2018powerful}).

\subsection{Proof of \cref{thm:nodeIDs}}


For the purposes of this proof, we assume the Universal Approximation Theorem \citep{hornik1989multilayer_universal} for multi-layer perceptrons (MLPs), stating that any continuous function with finite support can be approximated by an MLP with at least one hidden layer. We refer to this result to avoid giving explicit constructions for GNN layers. We also assume that the GNN uses min aggregation, consisting of an MLP applied elementwise to the set of incoming messages, followed by taking the minimum.

\begin{proof}
Initially, the ego node or \emph{root} node starts with the node feature $h^{(0)}(r) = 1$, which will be its final node ID. The other nodes, $v \in V(G) \setminus \{r\}$ start with $h^{(0)}(v) = 0$, which indicates that they have not yet been assigned a final ID. The node IDs will be assigned layer by layer, such that a node that is $k$ hops away from $r$ receives a final ID after $k$ rounds of message passing. 

We call a node \emph{active} if it has received a message from an active node in the previous round. At $t=0$, only the root node is active. Once a node is active, it switches to inactive for all the remaining rounds. A GNN can keep track of active nodes, by using a dimension of the hidden node embeddings to indicate whether the node is currently active and another dimension to indicate whether the node has already been active. This information is then ``included'' in its messages, so neighboring nodes know when to activate. Notice that only nodes that are exactly distance $k$ (where edges can be traversed in either direction) from the root node will be active in the $k^{th}$ round.

By ignoring messages from inactive nodes, a GNN can replicate \cref{alg:BFS_nodeID}. In each round, nodes receive messages from their neighbors, consisting of the send a message consisting of their ID and the incoming (or outgoing) port number to each of their incoming (or outgoing) neighbors respectively. The message is an integer in base $2n$ made by concatenating the node ID and the port number ($n$ is added to the port number if it is an incoming edge). The message can be viewed as a node ID proposal, every active node proposes a node ID to each of its neighbors. Then all nodes that receive proposals accept the smallest of these proposals (again in base $2n$) as their node ID and become active.

In a GNN, this can be replicated from the receiver's perspective. In each round, nodes receive messages from their neighbors, consisting of the neighboring node's current state and the port numbers along the edge. The MLP at each node $v$, ``ignores'' messages from inactive nodes, e.g. by encoding them as a very high value. For the remaining messages, $v$ encodes the neighbor's current state and port numbers as the concatenation of the neighbor's node ID (current state minus the two dimensions indicating activity) and the port number $v$ has at its neighbor. In this way, the MLP essentially constructs the proposals from the algorithm. Node $v$ then selects the minimum proposal.

Since our input graph is connected, every node in the graph will be reached (receive a proposal) within $D$ rounds, since $D$ is the diameter of the graph. Therefore every node will end up with a node ID. 

It remains to be shown that this node ID will be unique. First, note that nodes at different distances from the root cannot end up with the same node ID. A node at distance $k$ will receive its first proposal in round $k$ and will therefore have an ID with exactly $k+1$ digits. Finally, an inductive argument shows that active nodes (nodes at the same distance) cannot have the same node IDs. Certainly, this is true at the start when $X=\{r\}$. Now assuming all active nodes from the previous round ($k-1$) had distinct node IDs, then the only way two active nodes (in round $k$) can have the same ID is if they accept a proposal from the same neighboring node. This is because, assuming the induction hypothesis, proposals from different nodes will already differ in their first $k-1$ digits. But if two active nodes accepted a proposal from the same node, then they would have received different port numbers --- incoming ports are distinct, outgoing ports are distinct, and $n$ is added to all incoming ports so that they cannot be the same as any outgoing ports. Therefore the active nodes always accept unique proposals.
\end{proof}

The algorithm used in the proof is given in \cref{alg:BFS_nodeID}. \cref{fig:BFS_ex} gives an example output of the algorithm. 

\begin{algorithm}[tb]
\caption{BFS node ID assignment (\cref{thm:nodeIDs})}
\label{alg:BFS_nodeID}
\textbf{Input}: Connected directed multigraph $G=(V,E)$ with $n$ nodes, with diameter $D$ and with root (or ego) node $r \in V$. Active nodes $X \subseteq V$ and finished nodes $F \subseteq V$. Port numbering $P: E \rightarrow \mathbb{N}^{2}$.\\
\textbf{Output}: Unique node IDs $h(v)$ for all $v \in V$ (in base $2n$) 
\begin{algorithmic}[1]
\State $h(r) \gets 1$; \hspace{0.1in} $h(v) \gets 0$ for all $v \in V \setminus \{r\}$\;
\State $F \gets \emptyset$; \hspace{0.1in} $X \gets \{r\}$\;
\For{$k \gets 1$ to $D$}
\For{$v \in V$}
  \If {$v \in X$}
    \State send $h(v) \mathbin\Vert P((v,u))_{out}$ to $u \in N_{out}(v)$\;
    \State send $h(v) \mathbin\Vert n + P((u,v))_{in}$ to $u \in N_{in}(v)$\;
    \State $F \gets F \cup \{v\}$; \hspace{0.1in} $X \gets X \setminus \{v\}$\;
  \EndIf
  \If{$v \notin F$}
    \If{Incoming messages $M(v) \neq \emptyset$}
      \State $h(v) \gets \min\{M(v)\}$\;
      \State $X \gets X \cup \{v\}$\;
    \EndIf
  \EndIf
\EndFor
\EndFor
\end{algorithmic}
\end{algorithm}

\begin{figure}[ht]
\centering
\begin{tikzpicture}[scale=0.7, every node/.style={scale=0.9}, >=latex]
    \node[circle, draw, minimum size=0.4cm, fill=gray] (A) at  (-3.5,0) {};
    \node[circle, draw, minimum size=0.4cm] (B1) at  (0,2) {};
    \node[circle, draw, minimum size=0.4cm] (B2) at  (0,0) {};
    \node[circle, draw, minimum size=0.4cm] (B3) at  (0,-2) {};
    \node[circle, draw, minimum size=0.4cm] (C) at  (3.5,1) {};
    \draw [semithick,->] (B1) edge [bend right] node[pos=0.5,above] {(1,1)} (A);
    \draw [semithick,->] (A) edge node[pos=0.5,below] {(1,1)} (B1);
    \draw [semithick,->] (A) edge node[pos=0.5,below] {(1,2)} (B2);
    \draw [semithick,->] (B3) edge node[pos=0.5,below] {(2,1)} (A);
    \draw [semithick,->] (B1) edge node[pos=0.4,above right] {(1,2)} (C);
    \draw [semithick,->] (B2) edge node[pos=0.4,below right] {(2,1)} (C);
    \node at (A) [xshift=-0.2cm,anchor=east,text=blue] {$h = 1$};
    \node at (B1) [yshift=0.2cm,anchor=south,text=blue] {$h = 11, \text{\st{$16$}}$};
    \node at (B2) [yshift=0.2cm,anchor=south,text=blue] {$h = 12$};
    \node at (B3) [yshift=0.2cm,anchor=south,text=blue] {$h = 17$};
    \node at (C) [xshift=0.2cm,anchor=west,text=blue] {$h = 112, \text{\st{$121$}}$};
\end{tikzpicture}
\caption{Example output of Algorithm \ref{alg:BFS_nodeID} with the root (or ego) node indicated in gray. The edge labels show the (incoming, and outgoing) directed multigraph port numbers. Blue numbers indicate node ID proposals; declined proposals are struck through, leaving the final assigned node IDs. In this example $n=5$ so node IDs are to be understood in base $10$.}
\label{fig:BFS_ex}
\end{figure}
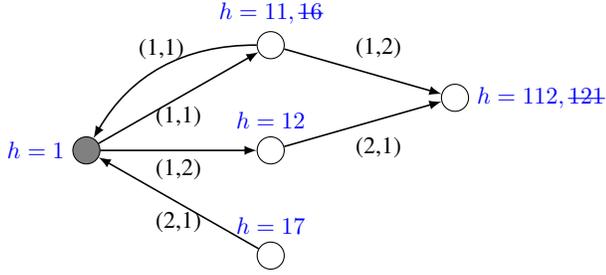

\section{Complexity \& Runtime}
\label{app:complexity}

We propose a set of adaptations, so the final time complexity of a model will depend on the choice of base GNN.
We describe the additional runtime costs incurred by the adaptations below. 
The empirical runtimes using GIN can be seen in \cref{tab:runtimes}, also in the appendix. 

Reverse message passing multiples the time complexity by a constant factor. Theoretically, this constant factor would be 2, but in practice, we also increase the memory requirements of the model by a factor of 2, so the batch size has to be reduced, leading to a larger constant factor (see \cref{tab:runtimes}).
Port numbering does not increase the time complexity of training and inference since it simply increases the size of the input edge features. This is confirmed by the empirical runtimes in \cref{tab:runtimes}. However, computing timestamp-based port numbers adds a pre-computation step of complexity $\mathcal{O}(m \log(m))$, which can be done in one go before training.
Ego IDs similarly do not increase the runtime complexity, as can be seen in \cref{tab:runtimes}.
All in all, the adaptations add a constant factor to the runtime complexity in addition to a pre-computation runtime of complexity $\mathcal{O}(m \log(m))$.

\section{Synthetic Pattern Detection Datasets}
\label{app:synth_pattern_detection}

To ensure that the desired subgraph patterns appear randomly, we introduce the \emph{random circulant graph} generator. The generator is parameterized by the number of nodes $n$, the average degree $d$, and a radius $r$ controlling the local density of the generated graph. Together, these parameters control the rate of generating the various subgraph patterns. For example, by choosing a high average degree and a low radius, we can ensure that more dense patterns also appear randomly. 

\subsection{Random Circulant Graphs}
\label{app:synth_graph_generator}

A \emph{random circulant graph}, $G_{n,d,r}$, has $n$ nodes $\{1,2,\ldots,n\}$, and $\floor{\frac{nd}{2}}$ edges; one end, $u$, of each edge is sampled uniformly from $\{1,2,\ldots,n\}$, and the other end, $v$, is sampled from a normal distribution around $u$ with standard deviation $r$, and then rounded to the nearest integer, i.e., $v = \floor{X + \frac{1}{2}}$, where $X \sim \mathcal{N}(u,r)$. \cref{alg:simulator} shows a step-by-step description of the algorithm. 

\begin{algorithm}[tb]
\caption{Random Circulant Graph Generator}
\label{alg:simulator}
\textbf{Parameters}: Number of nodes $n$, average degree $d$, locality radius $r$.\\
\textbf{Output}: A graph $G=(V,E)$.
\begin{algorithmic}[1]
\State $N_{edges} \gets \lfloor \frac{nd}{2} \rfloor$\;
\State $\Vec{u} \gets ( u_i \mid i = 1, \ldots, N_{edges} ) \overset{\text{iid}}{\sim} \text{unif}(\{1, \ldots, n\})$ \Comment{Sample tail nodes}
\State $\Vec{v} \gets ( v_i \mid i = 1, \ldots, N_{edges} ) \overset{\text{iid}}{\sim} \mathcal{N}(u_i, r)$ \Comment{Sample head nodes}
\State $\Vec{v} \gets ( \lfloor v_i + \frac{1}{2} \rfloor \mid i = 1, \ldots, N_{edges} )$ \Comment{Round head nodes to nearest integer}
\State $V(G) = \{1, \ldots, n \}$\;
\State $E(G) = \{ (u_i, v_i) \mid i = 1, \ldots, N_{edges} \}$\;
\end{algorithmic}
\end{algorithm}

This model is similar to the Watts–Strogatz small-world model \citep{watts1998collective}, where nodes are also arranged in a ring, each node is connected to its k nearest neighbors, and edges are randomly rewired with a chosen probability. However, standard Watts–Strogatz implementations produce undirected graphs and do not allow for multiple edges between a pair of vertices. In particular, a randomly oriented Watts-Strogatz graph --- a directed graph created by generating an undirected Watts-Strogatz graph, and then choosing an arbitrary direction for each edge independently and uniformly at random --- doesn't allow for testing the detection of directed cycles of length 2 and fan patterns
(without parallel edges fan patterns are identical to degree patterns).

\subsection{Pattern Detection Tasks}
\label{app:synth_tasks}

The synthetic pattern detection dataset includes the following (sub-)tasks:

\begin{itemize}
    \item \textbf{Degree-in (-out):} Degree -in and -out of a node is the number of incoming and outgoing transactions respectively. Note that multiple transactions can come from the same node, and these are counted separately.
    \item \textbf{Fan-in (-out):} Fan -in and -out of a node is the number of unique incoming and outgoing neighbors, respectively. Neighbors with multiple edges are counted only once, but a node can contribute to both the fan-in and fan-out counts, as it can be both an incoming and outgoing neighbor.
    \item \textbf{Cycle (k):} A node receives a positive label if it is part of a directed cycle of length $k$. We only consider directed cycle patterns and omit the word ``directed''.
    \item \textbf{Scatter-Gather:} A node receives a positive label if it is the \emph{sink} node of a scatter-gather pattern with at least $2$ \emph{intermediate} nodes. A scatter-gather pattern consists of a \emph{source} node, \emph{intermediate} nodes, and a \emph{sink} node. Given a \emph{sink} node $v$, a node $u$ is considered a \emph{source} node if there is a directed path of length $2$ from $u$ to $v$. A node $w_i$ is considered an \emph{intermediate} node if there is a directed path of length $2$ from $u$ to $v$ through $w_i$, i.e., if both $(u,w_i)$ and $(w_i,v)$ are in $E(G)$. Note that a scatter-gather pattern with a single intermediate node is simply a directed path of length $2$. This does not capture the pattern we are interested in, so we require at least $2$ intermediate nodes.
    \item \textbf{Gather-Scatter:} We do not explicitly include this pattern as a subtask since it is a combination of fan-in and fan-out.
    \item \textbf{Biclique:} A biclique (or complete bipartite graph) is a bipartite graph, where every node of the first set is connected to every node of the second set. We consider a special case of the directed biclique, where every node in the first set has at least one outgoing edge to every node in the second set. We call the nodes in the first set \emph{source} nodes, and the nodes in the second set \emph{sink} nodes. The task is to identify all nodes that are the sink nodes of some directed $K_{2,2}$ biclique with (at least) $2$ source nodes and (at least) $2$ sink nodes.
\end{itemize}

All the (sub-)tasks are combined in one overall task, i.e., the output of the models is a vector, where each element of the vector corresponds to one of the subtasks.

We implement pattern detection algorithms to find the desired patterns and generate binary node labels. The parameters are adjusted to achieve relatively balanced labels, but since a perfect split (for all subtasks) is not guaranteed, we rely on minority class F1 scores rather than accuracy to measure the performance of our models. All subtasks in the final dataset have at least $19\%$ and at most $68\%$ positive labels.

Thresholds are used in some cases to produce binary labels. For example, for the degree-in task, nodes with an in-degree greater than 3 receive the label $1$, while low in-degree nodes receive the label $0$. For degree-in/-out and fan-in/-out the threshold is 3. For degree-in-time-span and maximum-degree-in the thresholds are 1 and 4 respectively.
Examples of the patterns are provided in \cref{fig:aml_patterns}.

We use $n=8192$ nodes, average degree $d=6$, and radius $r=11.1$ to generate our pattern detection data. To avoid any information leakage between train, validation, and test sets, we generate an independent random circulant graph $G_{n,d,r}$ for each dataset.
\cref{table:sim_data} shows the thresholds and distribution of labels for each subtask. 

\begin{table}[t]
\centering
\begin{tabular}{lcc}
\toprule
\textbf{Subtask}            & Threshold     & Positive Ratio     \\ \midrule
Degree-in                   & >3            & 0.352              \\
Degree-out                  & >3            & 0.349              \\
Fan-in                      & >3            & 0.324              \\
Fan-out                     & >3            & 0.323              \\
Cycle (2)                   & -             & 0.191              \\
Cycle (3)                   & -             & 0.344              \\
Cycle (4)                   & -             & 0.527              \\
Cycle (5)                   & -             & 0.677              \\
Cycle (6)                   & -             & 0.779              \\
Scatter-Gather              & -             & 0.321              \\
Biclique                    & -             & 0.318              \\
\bottomrule
\end{tabular}
\caption{Synthetic pattern detection task statistics}
\label{table:sim_data}
\end{table}

\section{AML and ETH Datasets}
\label{app:aml_eth_data}

\begin{table}[t]
\centering
\begin{tabular}{lrrr}
\toprule
\textbf{Dataset}    & \# nodes      & \# edges      & Illicit Ratio \\ 
\midrule
AML Small HI        & 0.5M          & 5M            & 0.07\%        \\
AML Small LI        & 0.7M          & 7M            & 0.05\%        \\
AML Medium HI       & 2.1M          & 32M           & 0.11\%        \\
AML Medium LI       & 2.0M          & 31M           & 0.05\%        \\
\midrule
ETH                 & 2.9M          & 13M           & 0.04\%        \\
\bottomrule
\end{tabular}
\caption{AML and ETH task statistics. HI indicates a higher illicit ratio and LI indicates a lower illicit ratio.}
\label{table:aml_eth_data}
\end{table}

\subsection{AML Data Split}
\label{app:aml_split}

We use a 60-20-20 temporal train-validation-test split, i.e., we split the transaction indices after ordering them by their timestamps. The data split is defined by two timestamps $t_1$ and $t_2$.
Train indices correspond to transactions before time $t_1$, validation indices to transactions between times $t_1$ and $t_2$, and test indices to transactions after $t_2$. 
However, the validation and test set transactions need access to the previous transactions to identify patterns. So we construct train, validation, and test graphs.
This corresponds to considering the financial transaction graph as a dynamic graph and taking three snapshots at times $t_1$, $t_2$, and $t_3=t_{max}$.
The train graph contains only the training transactions (and corresponding nodes). The validation graph contains the training and validation transactions, but only the validation indices are used for evaluation. The test graph contains all the transactions, but only the test indices are used for evaluation. 
This is the most likely scenario faced by banks and financial authorities who want to classify new batches of transactions.

\subsection{ETH Dataset}
\label{app:eth_dataset}

The Kaggle dataset was constructed by starting from $1165$ reported phishing nodes and crawling the whole Ethereum transaction network via breadth-first search.
The dataset originally comes with four edge features: transaction time, transaction amount, source node address, and destination node address. Each node has one feature, the account address. As this information is minimal, we enhance the dataset with additional transaction features taken directly from the blockchain using Ethereum ETL \citep{ethetl}. We add nonce, block number, gas, and gas price. These are cryptocurrency terms for the number of transactions made by the sender prior to the given transaction; the block containing this transaction; and the transaction fees the sender provided for the transaction to go through. \cref{table:aml_eth_data} shows the number of nodes (accounts) and edges (transactions) and the illicit ratio. Unlike AML, this is a node classification task, where the goal is to classify nodes as phishing accounts.

Unlike the AML use case, we do not know a priori what subgraph patterns to look for to identify phishing nodes. To this end, we defined \emph{fraudulent clusters} and visualized these to give an indication of the patterns our model might need to detect. The clusters were constructed by starting from the $1165$ reported phishing nodes, adding their 1-hop neighborhoods, and then joining any clusters sharing a common node. \cref{fig:ethpatterns1} shows an example of one of these fraudulent clusters. 

\begin{figure}[ht]
    \centering
    \includegraphics[width=\columnwidth]{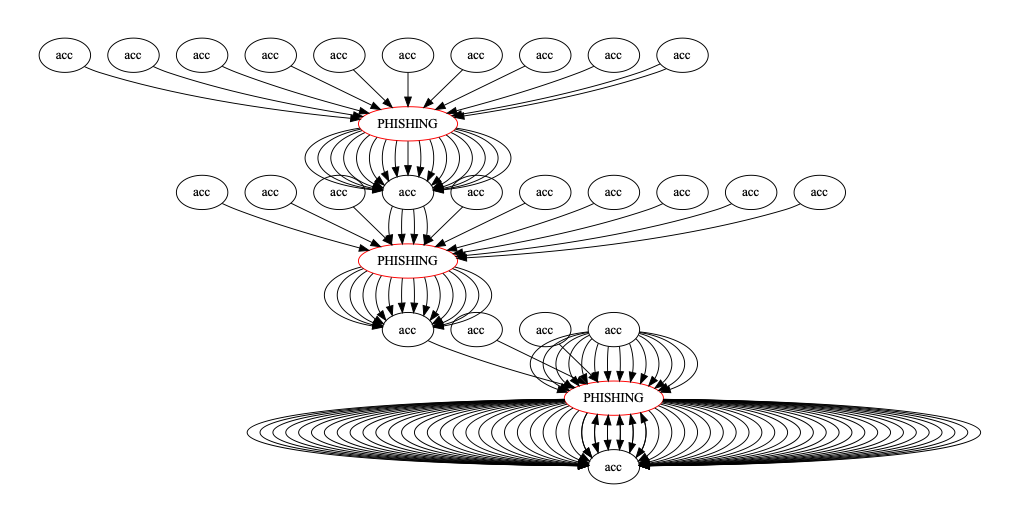}
    \caption{One of the fraudulent clusters identified in the ETH phishing dataset. The patterns that might help identify a fraudulent node are mostly 1-hop, namely in-degree, out-degree, fan-in, fan-out patterns, and 2-cycles.}
    \label{fig:ethpatterns1}
\end{figure}

Similar to AML, we use a temporal train-validation-test split, but this time we split the nodes. We order the nodes by the first transaction they are involved in (either as sender or receiver) before splitting.
Again this gives us threshold times $t_1$ and $t_2$, and we use these times to create our train, validation, and test graphs. See the AML dataset split for more details.  
This time we use a 65-15-20 split because the illicit accounts are skewed towards the end of the dataset, with few illicit accounts in the first half of the data.

\section{Experimental Details}
\label{app:exp_setup_details}

\subsection{Hyperparameter Tuning and Scoring}
\label{app:hyperparam_tuning}

We used random sampling to identify a good range of hyperparameters. A second round of random sampling was conducted with this narrower range to pick our final set of hyperparameters. We varied the following hyperparameters: the number of GNN layers, hidden embedding size, learning rate, dropout, and minority class weight (for the weighted loss function).
The number of random samples ranged between 10 to 50, depending on the training time of the model on a particular dataset.

To get our final results, we use the hyperparameters with the best validation score to train five models initialized with five different random seeds. The results of the five runs are then averaged and the standard deviation is calculated to give the reported scores.

\subsection{LightGBM with Graph-based Features}
\label{app:lightGBM}

We include a LightGBM baseline using pre-calculated graph features to classify nodes or edges individually.
Calculating complex subgraph features naively can be prohibitively expensive, therefore we rely on efficient implementations \citep{blanusa_fast_2023, blanusa2022scalable, altman2023realistic} to enumerate subgraph patterns. 

We use these graph-mining algorithms to enumerate the suspicious patterns introduced in \cref{fig:aml_patterns}. We enumerate on a fine-grained level, counting patterns of different sizes separately. For the edge classification tasks (AML) we count patterns that an edge is contained in and add these counts as additional edge features. We do the same with nodes for the node classification task (ETH). 

We train a LightGBM \citep{ke2017lightgbm} model on the individual edges (or nodes) with the full set of features (original raw features and additional graph-based features). 
The pattern enumeration can be restricted to a certain time range; we optimize the range (as a hyperparameter) for each dataset to get the best performance.

\subsection{Synthetic Pattern Detection Experiments}

Unlike the much larger AML and ETH datasets, we do not have to subsample the neighborhood. Instead, we use the whole 3-hop neighborhood in our model to ensure that all of the patterns can be found. Moreover, we used six GNN layers to ensure that 6-cycles can also be detected. 

\subsection{AML Experiments}

Since this is an edge classification dataset, all GNNs use a final edge readout layer that takes the edge embedding and the corresponding endpoint node embeddings as its input. We use neighborhood sampling, sampling $100$ one- and two-hop neighbors respectively.
We remove node IDs from edge and node features to avoid, as far as possible, overfitting to the fraudulent node IDs. 
Since we know what financial crime patterns were used for generating this dataset, we use the graph mining library to enumerate exactly these suspicious patterns for each transaction: degree-in/-out, fan-in/-out, directed cycle, undirected cycle, and scatter-gather patterns of different sizes. We leave out bicliques, as these are not currently included in the library.
For the low IR dataset, we took the hyperparameters from the corresponding high IR dataset and only optimized the minority class weight. We found that this has to be optimized to account for the greater class imbalance.

\subsection{AML Runtimes}
\label{app:runtimes}

\cref{tab:runtimes} shows the runtime cost of adding the adaptations to GIN when running on the AML Small HI dataset. The size of the GNN models was kept the same: 2 GNN layers and a hidden size of 64. All models were run on an Nvidia GeForce RTX 3090 GPU.

\begin{table}[t!]
    \centering
    \begin{tabular}{lrr}
\toprule
Model           & TTT (s) & TPS (inference) \\
\midrule
GIN                 &  1876 & 53130 \\
\midrule
\multicolumn{3}{l}{Individual Adaptations} \\
\midrule
GIN+Ports           &  2007 & 58663 \\
GIN+EgoIDs          &  1873 & 58533 \\
GIN+ReverseMP       & 12510 & 21067 \\
\midrule
\multicolumn{3}{l}{Cumulative Adaptations} \\
\midrule
GIN+ReverseMP+Ports & 13051 & 18763 \\
{\ \ +EgoIDs}       & 12958 & 18700 \\
\bottomrule
\end{tabular}
    \caption{Total training times (TTT) and Transaction per Second (TPS) for all GIN-based models on the AML Small HI dataset. The TPS values were measured in evaluation mode.}
    \label{tab:runtimes}
\end{table}

\subsection{ETH Experiments}

Since this is a node classification dataset, all GNNs use a final node readout layer that takes only the corresponding node embedding as its input. We use neighborhood sampling, sampling $1000$ one- and two-hop neighbors respectively.

For the LightGBM baseline, we use the graph mining library to enumerate the same graph-based patterns as for AML, but this time for nodes. Additionally, we calculate basic node statistics (such as averages, maximums, and variances) of the transaction amounts and the time between transactions to improve the baseline further. The statistical features give a significant boost to the minority class F1 scores, from $27.1\%$ with only raw and graph-based features to above $50\%$ with the additional statistical features.

\section{Additional Real-World Benchmarks}
\label{app:real_world_benchmarks}


The theory results and the subgraph detection tasks demonstrate the general purpose potential of the architectural adaptations. 
Indeed, many of the subgraph patterns we consider could also be relevant to other areas. 
However, testing our model on real-world benchmark datasets is crucial to support these claims. 
Therefore we have taken three real-world directed graph datasets and compared our approach with the state-of-the-art (SOTA) model for these 3 benchmarks \citep{rusch2022gradient}. A summary of the dataset statistics can be seen in \cref{table:real_data_stats}.

We report results for GIN with various adaptations. The models are set up as in our previous experiments. We use the Adam optimizer and train the model for 2000 epochs, using early stopping on the validation accuracy with a patience of 400. We do a single run of random sampling to optimize the hyperparameters within a predefined range. We sample $20$ different sets of hyperparameters, and take choose the best performing model. We use the standard data splits for the datasets: For Chameleon and Squirrel we use the fixed GEOM-GCN splits \citep{pei2020geomPlusSquirrel}, and for Arxiv-Year we use the splits provided by \citet{lim2021large}. We report the mean and standard deviation of the test accuracy, calculated over 10 different data splits.
The results can be seen in \cref{tab:new_datasets}. We outperform SOTA in 2 out of the 3 benchmarks. On one dataset in particular the gain is almost $6$ percentage points. 

\begin{table}[t]
    \centering
    \begingroup
    \resizebox{1.0\linewidth}{!}{%
    \begin{tabular}{lrrrr}
\toprule
\textbf{Dataset}    & \# nodes      & \# edges      & \# node feats & \# classes    \\ 
\midrule
Chamaleon           & 2,277         & 36K           & 2,325         & 5             \\
Squirrel            & 5,201         & 217K          & 2,089         & 5             \\
Arxiv-Year          & 169K          & 1.17M         & 128           & 40            \\
\bottomrule
\end{tabular}
    }
    \endgroup
    \caption{Real-World dataset statistics.}
    \label{table:real_data_stats}
\end{table}

\begin{table}
    \centering
    \begingroup
    \resizebox{1.0\linewidth}{!}{%
    \begin{tabular}{lcccc}
\toprule
{Model}             &               Squirrel &             Chameleon &            Arxiv-Year \\
\midrule
Grad. Gating (SOTA) &   $\bm{64.26 \pm 2.38}$&      $71.40 \pm 2.38$ &      $63.30 \pm 1.84$ \\
\midrule
GIN (GCN$^*$)       &     $35.94 \pm 2.01^*$ &      $51.29 \pm 0.63$ &      $50.15 \pm 0.26$ \\
{\ +ReverseMP}      &     $59.84 \pm 2.08^*$ &      $66.36 \pm 1.23$ &      $63.31 \pm 0.45$ \\
{\ +Ports}          &     $59.33 \pm 2.40^*$ &      $73.46 \pm 0.97$ &  $\bm{68.12 \pm 0.51}$\\
{\ +EgoIDs}         &     $58.41 \pm 2.49^*$ &  $\bm{73.73 \pm 1.03}$&      $68.04 \pm 0.60$ \\
\bottomrule
\end{tabular}

    }
    \endgroup
    \caption{Comparison with state-of-the-art \citep{rusch2022gradient} results on three real-world directed graph datasets. 
    }
    \label{tab:new_datasets}
\end{table}

\section{Additional Subgraph Detection Results}
\label{app:additional_experiments}

\begin{table}[t] 
    \centering
    \begingroup
    \setlength{\tabcolsep}{9pt}
    \resizebox{1.0\linewidth}{!}{%
    \begin{tabular}{lccccccccccc}
\toprule
Model 
& C4 & C5 & C6 & S-G & B-C \\
\midrule
DropGIN
& \cellcolor[HTML]{f7fcf5} \color[HTML]{000000} 16.92 & \cellcolor[HTML]{f7fcf5} \color[HTML]{000000} 26.02 & \cellcolor[HTML]{f7fcf5} \color[HTML]{000000} 45.17 & \cellcolor[HTML]{f7fcf5} \color[HTML]{000000} 48.87 & \cellcolor[HTML]{f7fcf5} \color[HTML]{000000} 51.59 \\
R-GCN
& \cellcolor[HTML]{e2f4dd} \color[HTML]{000000} 65.57 & \cellcolor[HTML]{e8f6e3} \color[HTML]{000000} 64.35 & \cellcolor[HTML]{bde5b6} \color[HTML]{000000} 71.27 & \cellcolor[HTML]{f7fcf5} \color[HTML]{000000} 59.23 & \cellcolor[HTML]{f7fcf5} \color[HTML]{000000} 55.61 \\
PPGN
& \cellcolor[HTML]{f7fcf5} \color[HTML]{000000} 57.93 & \cellcolor[HTML]{f7fcf5} \color[HTML]{000000} 48.74 & \cellcolor[HTML]{f7fcf5} \color[HTML]{000000} 41.15 & \cellcolor[HTML]{f7fcf5} \color[HTML]{000000} 54.23 & \cellcolor[HTML]{f7fcf5} \color[HTML]{000000} 54.98 \\
\midrule
Multi-DropGIN
& \cellcolor[HTML]{004a1e} \color[HTML]{f1f1f1} 99.09 & \cellcolor[HTML]{005b25} \color[HTML]{f1f1f1} 97.08 & \cellcolor[HTML]{006d2c} \color[HTML]{f1f1f1} 94.93 & \cellcolor[HTML]{005c25} \color[HTML]{f1f1f1} 96.98 & \cellcolor[HTML]{05712f} \color[HTML]{f1f1f1} 94.27 \\
Multi-R-GCN
& \cellcolor[HTML]{3fa95c} \color[HTML]{f1f1f1} 85.20 & \cellcolor[HTML]{a8dca2} \color[HTML]{000000} 74.08 & \cellcolor[HTML]{a7dba0} \color[HTML]{000000} 74.30 & \cellcolor[HTML]{5bb86a} \color[HTML]{f1f1f1} 82.37 & \cellcolor[HTML]{d9f0d3} \color[HTML]{000000} 67.16 \\
Multi-PPGN$^*$
& \cellcolor[HTML]{005321} \color[HTML]{f1f1f1} 98.11 & \cellcolor[HTML]{127c39} \color[HTML]{f1f1f1} 92.49 & \cellcolor[HTML]{37a055} \color[HTML]{f1f1f1} 86.69 & \cellcolor[HTML]{cfecc9} \color[HTML]{000000} 68.74 & \cellcolor[HTML]{edf8ea} \color[HTML]{000000} 62.78 \\
\midrule
Multi-PNA 
& \cellcolor[HTML]{00481d} \color[HTML]{f1f1f1} 99.49 & \cellcolor[HTML]{005924} \color[HTML]{f1f1f1} 97.46 & \cellcolor[HTML]{2a924a} \color[HTML]{f1f1f1} 88.75 & \cellcolor[HTML]{004a1e} \color[HTML]{f1f1f1} 99.07 & \cellcolor[HTML]{005e26} \color[HTML]{f1f1f1} 96.77 \\
\bottomrule
\end{tabular}
    }
    \endgroup
    \caption{Additional subgraph detection results (F1 scores). Multi-PNA is included for comparison. $^*$No port numbers. 
    }
    \label{tab:new_sim_results}
\end{table}

\begin{table*}
    \centering
    \begin{tabular}{rccccccc}
\toprule
{No. Training Nodes} & {C4} & {C5} & {C6} & {S-G} & {B-C} \\
\midrule
1024 & 
\cellcolor[HTML]{e4f5df} \color[HTML]{000000} 65.22 $\pm$ 3.34 & \cellcolor[HTML]{f2faf0} \color[HTML]{000000} 61.38 $\pm$ 7.25 & \cellcolor[HTML]{f7fcf5} \color[HTML]{000000} 59.97 $\pm$ 7.30 & \cellcolor[HTML]{e5f5e0} \color[HTML]{000000} 65.12 $\pm$ 2.83 & \cellcolor[HTML]{f7fcf5} \color[HTML]{000000} 60.11 $\pm$ 2.86 \\
2048 & 
\cellcolor[HTML]{86cc85} \color[HTML]{000000} 78.06 $\pm$ 6.19 & \cellcolor[HTML]{b5e1ae} \color[HTML]{000000} 72.44 $\pm$ 4.48 & \cellcolor[HTML]{c6e8bf} \color[HTML]{000000} 70.19 $\pm$ 2.49 & \cellcolor[HTML]{99d595} \color[HTML]{000000} 75.82 $\pm$ 4.77 & \cellcolor[HTML]{d4eece} \color[HTML]{000000} 67.82 $\pm$ 1.53 \\
4096 & 
\cellcolor[HTML]{006529} \color[HTML]{f1f1f1} 95.85 $\pm$ 0.86 & \cellcolor[HTML]{147e3a} \color[HTML]{f1f1f1} 92.06 $\pm$ 1.81 & \cellcolor[HTML]{92d28f} \color[HTML]{000000} 76.57 $\pm$ 0.96 & \cellcolor[HTML]{0e7936} \color[HTML]{f1f1f1} 92.90 $\pm$ 1.29 & \cellcolor[HTML]{258d47} \color[HTML]{f1f1f1} 89.64 $\pm$ 1.74 \\
8192$^*$ & 
\cellcolor[HTML]{005924} \color[HTML]{f1f1f1} 97.46 $\pm$ 0.43 & \cellcolor[HTML]{17813d} \color[HTML]{f1f1f1} 91.60 $\pm$ 1.50 & \cellcolor[HTML]{48ae60} \color[HTML]{f1f1f1} 84.23 $\pm$ 1.12 & \cellcolor[HTML]{005924} \color[HTML]{f1f1f1} 97.42 $\pm$ 0.85 & \cellcolor[HTML]{05712f} \color[HTML]{f1f1f1} 94.33 $\pm$ 0.95 \\
8192 & 
\cellcolor[HTML]{00481d} \color[HTML]{f1f1f1} 99.46 $\pm$ 0.05 & \cellcolor[HTML]{005522} \color[HTML]{f1f1f1} 97.92 $\pm$ 0.26 & \cellcolor[HTML]{097532} \color[HTML]{f1f1f1} 93.72 $\pm$ 0.75 & \cellcolor[HTML]{004d1f} \color[HTML]{f1f1f1} 98.77 $\pm$ 0.18 & \cellcolor[HTML]{005622} \color[HTML]{f1f1f1} 97.75 $\pm$ 0.21 \\
16384 & 
\cellcolor[HTML]{00451c} \color[HTML]{f1f1f1} 99.72 $\pm$ 0.06 & \cellcolor[HTML]{004e1f} \color[HTML]{f1f1f1} 98.66 $\pm$ 0.15 & \cellcolor[HTML]{005a24} \color[HTML]{f1f1f1} 97.24 $\pm$ 0.19 & \cellcolor[HTML]{00471c} \color[HTML]{f1f1f1} 99.61 $\pm$ 0.08 & \cellcolor[HTML]{00481d} \color[HTML]{f1f1f1} 99.53 $\pm$ 0.07 \\
\bottomrule
\end{tabular}
    \caption{Minority class F1 scores for the ``complex'' subgraph detection tasks with different dataset sizes. Multi-GIN is used in all experiments. The $^*$ indicates that these results are copied from \cref{tab:sim_results}, where all subgraph tasks were learned at once. All other experiments were restricted to the subtasks shown. 
    }
    \label{tab:sim_size_results}
\end{table*}

\begin{table*}
    \centering
    \begin{tabular}{rccccccccc}
\toprule
{No. Training Nodes} & {C2} & {C3} & {C4} & {C5} & {C6} & {C7} & {C8} & {C9} & {C10} \\
\midrule

2048 & 
\cellcolor[HTML]{006c2c} \color[HTML]{f1f1f1} 97.53 & \cellcolor[HTML]{79c67a} \color[HTML]{000000} 89.72 & \cellcolor[HTML]{cbeac4} \color[HTML]{000000} 84.71 & \cellcolor[HTML]{f3faf0} \color[HTML]{000000} 80.60 & \cellcolor[HTML]{f7fcf5} \color[HTML]{000000} 75.48 & \cellcolor[HTML]{edf8ea} \color[HTML]{000000} 81.34 & \cellcolor[HTML]{55b567} \color[HTML]{f1f1f1} 91.51 & \cellcolor[HTML]{349d53} \color[HTML]{f1f1f1} 93.57 & \cellcolor[HTML]{369f54} \color[HTML]{f1f1f1} 93.39 \\
4096 & 
\cellcolor[HTML]{004c1e} \color[HTML]{f1f1f1} 99.51 & \cellcolor[HTML]{0d7836} \color[HTML]{f1f1f1} 96.49 & \cellcolor[HTML]{208843} \color[HTML]{f1f1f1} 95.23 & \cellcolor[HTML]{79c67a} \color[HTML]{000000} 89.75 & \cellcolor[HTML]{bce4b5} \color[HTML]{000000} 85.71 & \cellcolor[HTML]{bbe4b4} \color[HTML]{000000} 85.85 & \cellcolor[HTML]{66bd6f} \color[HTML]{f1f1f1} 90.69 & \cellcolor[HTML]{29914a} \color[HTML]{f1f1f1} 94.52 & \cellcolor[HTML]{147e3a} \color[HTML]{f1f1f1} 96.09 \\
8192 & 
\cellcolor[HTML]{00471c} \color[HTML]{f1f1f1} 99.83 & \cellcolor[HTML]{00471c} \color[HTML]{f1f1f1} 99.78 & \cellcolor[HTML]{005221} \color[HTML]{f1f1f1} 99.13 & \cellcolor[HTML]{05712f} \color[HTML]{f1f1f1} 97.17 & \cellcolor[HTML]{349d53} \color[HTML]{f1f1f1} 93.58 & \cellcolor[HTML]{75c477} \color[HTML]{000000} 89.93 & \cellcolor[HTML]{4eb264} \color[HTML]{f1f1f1} 91.87 & \cellcolor[HTML]{248c46} \color[HTML]{f1f1f1} 94.92 & \cellcolor[HTML]{0c7735} \color[HTML]{f1f1f1} 96.64 \\
16384 & 
\cellcolor[HTML]{00441b} \color[HTML]{f1f1f1} 99.96 & \cellcolor[HTML]{00441b} \color[HTML]{f1f1f1} 99.95 & \cellcolor[HTML]{00481d} \color[HTML]{f1f1f1} 99.75 & \cellcolor[HTML]{005321} \color[HTML]{f1f1f1} 99.01 & \cellcolor[HTML]{006d2c} \color[HTML]{f1f1f1} 97.46 & \cellcolor[HTML]{16803c} \color[HTML]{f1f1f1} 95.89 & \cellcolor[HTML]{218944} \color[HTML]{f1f1f1} 95.15 & \cellcolor[HTML]{0e7936} \color[HTML]{f1f1f1} 96.45 & \cellcolor[HTML]{006b2b} \color[HTML]{f1f1f1} 97.64 \\
32768 & 
\cellcolor[HTML]{00441b} \color[HTML]{f1f1f1} 99.99 & \cellcolor[HTML]{00441b} \color[HTML]{f1f1f1} 100.00 & \cellcolor[HTML]{00441b} \color[HTML]{f1f1f1} 99.93 & \cellcolor[HTML]{00491d} \color[HTML]{f1f1f1} 99.62 & \cellcolor[HTML]{005723} \color[HTML]{f1f1f1} 98.76 & \cellcolor[HTML]{006428} \color[HTML]{f1f1f1} 97.98 & \cellcolor[HTML]{006529} \color[HTML]{f1f1f1} 97.93 & \cellcolor[HTML]{006328} \color[HTML]{f1f1f1} 98.10 & \cellcolor[HTML]{005f26} \color[HTML]{f1f1f1} 98.30 \\
65536 & 
\cellcolor[HTML]{00441b} \color[HTML]{f1f1f1} 100.00 & \cellcolor[HTML]{00441b} \color[HTML]{f1f1f1} 100.00 & \cellcolor[HTML]{00441b} \color[HTML]{f1f1f1} 99.96 & \cellcolor[HTML]{00471c} \color[HTML]{f1f1f1} 99.80 & \cellcolor[HTML]{004e1f} \color[HTML]{f1f1f1} 99.36 & \cellcolor[HTML]{005622} \color[HTML]{f1f1f1} 98.89 & \cellcolor[HTML]{005924} \color[HTML]{f1f1f1} 98.73 & \cellcolor[HTML]{005622} \color[HTML]{f1f1f1} 98.86 & \cellcolor[HTML]{005622} \color[HTML]{f1f1f1} 98.83 \\
\bottomrule
\end{tabular}
    \caption{
    Minority class F1 scores for detecting cycles of increasing lengths with different training dataset sizes. Multi-GIN is used in all experiments. 
    }
    \label{tab:sim_cycles}
\end{table*}

\begin{figure}
    \centering
    \includegraphics[width=1.0\linewidth]{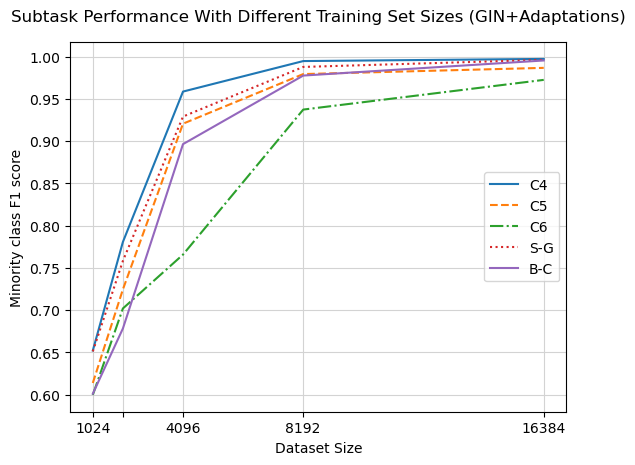}
    \caption{Performance of Multi-GIN on the ``complex'' subtasks as training dataset size is increased.}
    \label{fig:sim_size_results}
\end{figure}

\begin{table*}
    \centering
    \begingroup
\resizebox{\linewidth}{!}{%
\begin{tabular}{lccccccccccc}
\toprule
Model & deg-in & deg-out & fan-in & fan-out & C2 & C3 & C4 & C5 & C6 & S-G & B-C \\
\midrule
GIN & 
\cellcolor[HTML]{00471c} \color[HTML]{f1f1f1} 99.63 & \cellcolor[HTML]{f7fcf5} \color[HTML]{000000} 45.70 & \cellcolor[HTML]{00481d} \color[HTML]{f1f1f1} 99.49 & \cellcolor[HTML]{f7fcf5} \color[HTML]{000000} 42.15 & \cellcolor[HTML]{f7fcf5} \color[HTML]{000000} 35.48 & \cellcolor[HTML]{f7fcf5} \color[HTML]{000000} 57.04 & \cellcolor[HTML]{f7fcf5} \color[HTML]{000000} 51.07 & \cellcolor[HTML]{f7fcf5} \color[HTML]{000000} 45.66 & \cellcolor[HTML]{f7fcf5} \color[HTML]{000000} 48.32 & \cellcolor[HTML]{cbeac4} \color[HTML]{000000} 69.38 & \cellcolor[HTML]{e9f7e5} \color[HTML]{000000} 63.85 \\
GIN+EgoIDs & 
\cellcolor[HTML]{00441b} \color[HTML]{f1f1f1} 99.90 & \cellcolor[HTML]{f7fcf5} \color[HTML]{000000} 51.51 & \cellcolor[HTML]{00471c} \color[HTML]{f1f1f1} 99.64 & \cellcolor[HTML]{f7fcf5} \color[HTML]{000000} 49.92 & \cellcolor[HTML]{004d1f} \color[HTML]{f1f1f1} 98.89 & \cellcolor[HTML]{097532} \color[HTML]{f1f1f1} 93.71 & \cellcolor[HTML]{f7fcf5} \color[HTML]{000000} 57.89 & \cellcolor[HTML]{f7fcf5} \color[HTML]{000000} 49.08 & \cellcolor[HTML]{f7fcf5} \color[HTML]{000000} 49.78 & \cellcolor[HTML]{d8f0d2} \color[HTML]{000000} 67.27 & \cellcolor[HTML]{edf8e9} \color[HTML]{000000} 62.88 \\
GIN+Ports & 
\cellcolor[HTML]{00451c} \color[HTML]{f1f1f1} 99.84 & \cellcolor[HTML]{f7fcf5} \color[HTML]{000000} 42.33 & \cellcolor[HTML]{00481d} \color[HTML]{f1f1f1} 99.51 & \cellcolor[HTML]{f7fcf5} \color[HTML]{000000} 38.61 & \cellcolor[HTML]{f7fcf5} \color[HTML]{000000} 33.72 & \cellcolor[HTML]{f7fcf5} \color[HTML]{000000} 55.77 & \cellcolor[HTML]{f7fcf5} \color[HTML]{000000} 55.43 & \cellcolor[HTML]{f7fcf5} \color[HTML]{000000} 47.25 & \cellcolor[HTML]{f7fcf5} \color[HTML]{000000} 46.79 & \cellcolor[HTML]{ceecc8} \color[HTML]{000000} 68.89 & \cellcolor[HTML]{e1f3dc} \color[HTML]{000000} 65.68 \\
GIN+ReverseMP & 
\cellcolor[HTML]{004e1f} \color[HTML]{f1f1f1} 98.64 & \cellcolor[HTML]{004e1f} \color[HTML]{f1f1f1} 98.73 & \cellcolor[HTML]{005622} \color[HTML]{f1f1f1} 97.74 & \cellcolor[HTML]{004d1f} \color[HTML]{f1f1f1} 98.83 & \cellcolor[HTML]{f7fcf5} \color[HTML]{000000} 40.09 & \cellcolor[HTML]{f3faf0} \color[HTML]{000000} 61.10 & \cellcolor[HTML]{b6e2af} \color[HTML]{000000} 72.34 & \cellcolor[HTML]{c6e8bf} \color[HTML]{000000} 70.27 & \cellcolor[HTML]{a4da9e} \color[HTML]{000000} 74.55 & \cellcolor[HTML]{dcf2d7} \color[HTML]{000000} 66.44 & \cellcolor[HTML]{ddf2d8} \color[HTML]{000000} 66.26 \\
{\ +Ports} & 
\cellcolor[HTML]{004c1e} \color[HTML]{f1f1f1} 99.03 & \cellcolor[HTML]{00481d} \color[HTML]{f1f1f1} 99.44 & \cellcolor[HTML]{00491d} \color[HTML]{f1f1f1} 99.35 & \cellcolor[HTML]{00491d} \color[HTML]{f1f1f1} 99.37 & \cellcolor[HTML]{f7fcf5} \color[HTML]{000000} 43.15 & \cellcolor[HTML]{ecf8e8} \color[HTML]{000000} 63.15 & \cellcolor[HTML]{d2edcc} \color[HTML]{000000} 68.20 & \cellcolor[HTML]{d1edcb} \color[HTML]{000000} 68.29 & \cellcolor[HTML]{abdda5} \color[HTML]{000000} 73.67 & \cellcolor[HTML]{d3eecd} \color[HTML]{000000} 68.02 & \cellcolor[HTML]{dbf1d5} \color[HTML]{000000} 66.87 \\
{\ +EgoIDs (Multi-GIN)} & 
\cellcolor[HTML]{00471c} \color[HTML]{f1f1f1} 99.68 & \cellcolor[HTML]{00451c} \color[HTML]{f1f1f1} 99.71 & \cellcolor[HTML]{00451c} \color[HTML]{f1f1f1} 99.76 & \cellcolor[HTML]{00481d} \color[HTML]{f1f1f1} 99.52 & \cellcolor[HTML]{004c1e} \color[HTML]{f1f1f1} 99.02 & \cellcolor[HTML]{004a1e} \color[HTML]{f1f1f1} 99.13 & \cellcolor[HTML]{005522} \color[HTML]{f1f1f1} 97.93 & \cellcolor[HTML]{087432} \color[HTML]{f1f1f1} 93.88 & \cellcolor[HTML]{56b567} \color[HTML]{f1f1f1} 82.83 & \cellcolor[HTML]{00481d} \color[HTML]{f1f1f1} 99.47 & \cellcolor[HTML]{004d1f} \color[HTML]{f1f1f1} 98.81 \\

\bottomrule
\end{tabular}
}
\endgroup
    \caption{Subgraph detection results with parallel edges combined into single weighted edges. 
    }
    \label{tab:sim_collapsed}
\end{table*}

\begin{table*}
    \centering
    \begingroup
\resizebox{\linewidth}{!}{%
\begin{tabular}{lccccccccccc}
\toprule
Model & deg-in & deg-out & fan-in & fan-out & C2 & C3 & C4 & C5 & C6 & S-G & B-C \\
\midrule
GIN & 
\cellcolor[HTML]{00441b} \color[HTML]{f1f1f1} 99.91 & \cellcolor[HTML]{f7fcf5} \color[HTML]{000000} 44.68 & \cellcolor[HTML]{006b2b} \color[HTML]{f1f1f1} 95.26 & \cellcolor[HTML]{f7fcf5} \color[HTML]{000000} 40.72 & \cellcolor[HTML]{f7fcf5} \color[HTML]{000000} 25.05 & \cellcolor[HTML]{f7fcf5} \color[HTML]{000000} 55.30 & \cellcolor[HTML]{f7fcf5} \color[HTML]{000000} 46.73 & \cellcolor[HTML]{f7fcf5} \color[HTML]{000000} 43.61 & \cellcolor[HTML]{f7fcf5} \color[HTML]{000000} 47.32 & \cellcolor[HTML]{dff3da} \color[HTML]{000000} 65.97 & \cellcolor[HTML]{ebf7e7} \color[HTML]{000000} 63.58 \\
GIN+EgoIDs & 
\cellcolor[HTML]{00451c} \color[HTML]{f1f1f1} 99.82 & \cellcolor[HTML]{f7fcf5} \color[HTML]{000000} 52.56 & \cellcolor[HTML]{006729} \color[HTML]{f1f1f1} 95.67 & \cellcolor[HTML]{f7fcf5} \color[HTML]{000000} 49.78 & \cellcolor[HTML]{004e1f} \color[HTML]{f1f1f1} 98.67 & \cellcolor[HTML]{006729} \color[HTML]{f1f1f1} 95.78 & \cellcolor[HTML]{f7fcf5} \color[HTML]{000000} 54.56 & \cellcolor[HTML]{f7fcf5} \color[HTML]{000000} 45.95 & \cellcolor[HTML]{f7fcf5} \color[HTML]{000000} 48.01 & \cellcolor[HTML]{cdecc7} \color[HTML]{000000} 69.00 & \cellcolor[HTML]{e5f5e1} \color[HTML]{000000} 64.93 \\
GIN+Ports & 
\cellcolor[HTML]{004d1f} \color[HTML]{f1f1f1} 98.88 & \cellcolor[HTML]{f7fcf5} \color[HTML]{000000} 44.49 & \cellcolor[HTML]{004a1e} \color[HTML]{f1f1f1} 99.16 & \cellcolor[HTML]{f7fcf5} \color[HTML]{000000} 40.71 & \cellcolor[HTML]{f7fcf5} \color[HTML]{000000} 27.84 & \cellcolor[HTML]{f7fcf5} \color[HTML]{000000} 55.73 & \cellcolor[HTML]{f7fcf5} \color[HTML]{000000} 48.03 & \cellcolor[HTML]{f7fcf5} \color[HTML]{000000} 43.30 & \cellcolor[HTML]{f7fcf5} \color[HTML]{000000} 47.78 & \cellcolor[HTML]{dcf2d7} \color[HTML]{000000} 66.46 & \cellcolor[HTML]{ecf8e8} \color[HTML]{000000} 63.04 \\
GIN+ReverseMP & 
\cellcolor[HTML]{005020} \color[HTML]{f1f1f1} 98.58 & \cellcolor[HTML]{00491d} \color[HTML]{f1f1f1} 99.24 & \cellcolor[HTML]{0c7735} \color[HTML]{f1f1f1} 93.19 & \cellcolor[HTML]{05712f} \color[HTML]{f1f1f1} 94.23 & \cellcolor[HTML]{f7fcf5} \color[HTML]{000000} 38.18 & \cellcolor[HTML]{eef8ea} \color[HTML]{000000} 62.62 & \cellcolor[HTML]{c9eac2} \color[HTML]{000000} 69.74 & \cellcolor[HTML]{ccebc6} \color[HTML]{000000} 69.08 & \cellcolor[HTML]{a8dca2} \color[HTML]{000000} 74.07 & \cellcolor[HTML]{e0f3db} \color[HTML]{000000} 65.83 & \cellcolor[HTML]{eaf7e6} \color[HTML]{000000} 63.67 \\
{\ +Ports} & 
\cellcolor[HTML]{005120} \color[HTML]{f1f1f1} 98.36 & \cellcolor[HTML]{005120} \color[HTML]{f1f1f1} 98.31 & \cellcolor[HTML]{004c1e} \color[HTML]{f1f1f1} 99.01 & \cellcolor[HTML]{00491d} \color[HTML]{f1f1f1} 99.37 & \cellcolor[HTML]{f7fcf5} \color[HTML]{000000} 37.13 & \cellcolor[HTML]{f0f9ed} \color[HTML]{000000} 61.88 & \cellcolor[HTML]{b7e2b1} \color[HTML]{000000} 72.12 & \cellcolor[HTML]{c0e6b9} \color[HTML]{000000} 71.04 & \cellcolor[HTML]{94d390} \color[HTML]{000000} 76.47 & \cellcolor[HTML]{e5f5e0} \color[HTML]{000000} 65.00 & \cellcolor[HTML]{daf0d4} \color[HTML]{000000} 66.89 \\
{\ +EgoIDs (Multi-GIN)} & 
\cellcolor[HTML]{00491d} \color[HTML]{f1f1f1} 99.31 & \cellcolor[HTML]{004c1e} \color[HTML]{f1f1f1} 99.05 & \cellcolor[HTML]{00491d} \color[HTML]{f1f1f1} 99.36 & \cellcolor[HTML]{00491d} \color[HTML]{f1f1f1} 99.37 & \cellcolor[HTML]{005221} \color[HTML]{f1f1f1} 98.28 & \cellcolor[HTML]{005321} \color[HTML]{f1f1f1} 97.97 & \cellcolor[HTML]{006529} \color[HTML]{f1f1f1} 95.92 & \cellcolor[HTML]{2f984f} \color[HTML]{f1f1f1} 87.82 & \cellcolor[HTML]{84cc83} \color[HTML]{000000} 78.15 & \cellcolor[HTML]{005e26} \color[HTML]{f1f1f1} 96.83 & \cellcolor[HTML]{19833e} \color[HTML]{f1f1f1} 91.26 \\
\bottomrule
\end{tabular}
}
\endgroup

    \caption{Subgraph detection results using random unique integer node IDs. 
    }
    \label{tab:sim_node_ids}
\end{table*}

\begin{table*}[t] 
    \centering
    \begingroup
    \setlength{\tabcolsep}{4pt}
    \begin{tabular}{lccccccc}
\toprule
Model 
& Fan-in 
& Fan-out 
& Cycle 
& S-G 
& G-S 
& B-C 
& none
\\
\midrule
\rule{1cm}{0pt}&\rule{1cm}{0pt}&\rule{1cm}{0pt}&\rule{1cm}{0pt}&\rule{1cm}{0pt}&\rule{1cm}{0pt}&\rule{1cm}{0pt}&\rule{1cm}{0pt}\\[-\arraystretch\normalbaselineskip]

GIN \citep{xu2018powerful, hu2019strategies} 
& \cellcolor[HTML]{95d391} \color[HTML]{000000} 40.65 
& \cellcolor[HTML]{3ba458} \color[HTML]{f1f1f1} 64.89 
& \cellcolor[HTML]{88ce87} \color[HTML]{000000} 44.44 
& \cellcolor[HTML]{6ec173} \color[HTML]{000000} 51.46 
& \cellcolor[HTML]{7fc97f} \color[HTML]{000000} 47.24 
& \cellcolor[HTML]{a9dca3} \color[HTML]{000000} 34.85 
& \cellcolor[HTML]{f1faee} \color[HTML]{000000} 4.31
\\



GIN+ReverseMP \citep{jaume2019edgnnbidirecibmmaria}
& \cellcolor[HTML]{52b365} \color[HTML]{f1f1f1} 58.54 
& \cellcolor[HTML]{127c39} \color[HTML]{f1f1f1} 80.92 
& \cellcolor[HTML]{97d492} \color[HTML]{000000} 40.40 
& \cellcolor[HTML]{289049} \color[HTML]{f1f1f1} 72.80 
& \cellcolor[HTML]{329b51} \color[HTML]{f1f1f1} 68.50 
& \cellcolor[HTML]{a4da9e} \color[HTML]{000000} 36.36 
& \cellcolor[HTML]{f4fbf1} \color[HTML]{000000} 2.43
\\
{\ +Ports} 
& \cellcolor[HTML]{359e53} \color[HTML]{f1f1f1} 67.48 
& \cellcolor[HTML]{005f26} \color[HTML]{f1f1f1} 91.60 
& \cellcolor[HTML]{79c67a} \color[HTML]{000000} 48.48 
& \cellcolor[HTML]{0c7735} \color[HTML]{f1f1f1} 82.85 
& \cellcolor[HTML]{087432} \color[HTML]{f1f1f1} 84.51 
& \cellcolor[HTML]{6ec173} \color[HTML]{000000} 51.52 
& \cellcolor[HTML]{f6fcf4} \color[HTML]{000000} 1.08
\\
{\ +EgoIDs (Multi-GIN)}
& \cellcolor[HTML]{006c2c} \color[HTML]{f1f1f1} 87.80 
& \cellcolor[HTML]{147e3a} \color[HTML]{f1f1f1} 80.15 
& \cellcolor[HTML]{72c375} \color[HTML]{000000} 50.51 
& \cellcolor[HTML]{006428} \color[HTML]{f1f1f1} 89.96 
& \cellcolor[HTML]{026f2e} \color[HTML]{f1f1f1} 86.35 
& \cellcolor[HTML]{6ec173} \color[HTML]{000000} 51.52 
& \cellcolor[HTML]{f5fbf2} \color[HTML]{000000} 1.89
\\

\midrule

Multi-GIN+EU
& \cellcolor[HTML]{00692a} \color[HTML]{f1f1f1} 88.62 
& \cellcolor[HTML]{005221} \color[HTML]{f1f1f1} 95.42 
& \cellcolor[HTML]{55b567} \color[HTML]{f1f1f1} 57.58 
& \cellcolor[HTML]{00692a} \color[HTML]{f1f1f1} 88.28 
& \cellcolor[HTML]{005622} \color[HTML]{f1f1f1} 94.49 
& \cellcolor[HTML]{55b567} \color[HTML]{f1f1f1} 57.58 
& \cellcolor[HTML]{f5fbf3} \color[HTML]{000000} 1.35
\\
Multi-PNA
& \cellcolor[HTML]{067230} \color[HTML]{f1f1f1} 85.37 
& \cellcolor[HTML]{005a24} \color[HTML]{f1f1f1} 93.13 
& \cellcolor[HTML]{45ad5f} \color[HTML]{f1f1f1} 61.62 
& \cellcolor[HTML]{005924} \color[HTML]{f1f1f1} 93.72 
& \cellcolor[HTML]{005b25} \color[HTML]{f1f1f1} 92.65 
& \cellcolor[HTML]{62bb6d} \color[HTML]{f1f1f1} 54.55 
& \cellcolor[HTML]{f6fcf4} \color[HTML]{000000} 0.54
\\

\bottomrule
\end{tabular}
    \endgroup
    \caption{Recall scores by money laundering pattern on AML Small HI dataset, where ground truth pattern labels are available.
    S-G stands for Scatter-Gather, G-S stands for Gather-Scatter, B-C stands for Bipartite, and none indicated illicit transactions that are not part of any standard money laundering pattern. The ground truth money laundering patterns are of varying sizes. 
    }
    \label{tab:aml_small_hi_patterns}
\end{table*}

\begin{table*}[t] 
    \centering
    \begingroup
    \setlength{\tabcolsep}{4pt}
    \begin{tabular}{lccccccc}
\toprule
Model 
& Fan-in 
& Fan-out 
& Cycle 
& S-G 
& G-S 
& B-C 
& none
\\
\midrule
\rule{1cm}{0pt}&\rule{1cm}{0pt}&\rule{1cm}{0pt}&\rule{1cm}{0pt}&\rule{1cm}{0pt}&\rule{1cm}{0pt}&\rule{1cm}{0pt}&\rule{1cm}{0pt}\\[-\arraystretch\normalbaselineskip]

GIN \citep{xu2018powerful, hu2019strategies} 
& \cellcolor[HTML]{f7fcf5} \color[HTML]{000000} 0.00 
& \cellcolor[HTML]{48ae60} \color[HTML]{f1f1f1} 60.87 
& \cellcolor[HTML]{e6f5e1} \color[HTML]{000000} 11.90 
& \cellcolor[HTML]{d2edcc} \color[HTML]{000000} 20.34 
& \cellcolor[HTML]{92d28f} \color[HTML]{000000} 41.43 
& \cellcolor[HTML]{f4fbf1} \color[HTML]{000000} 2.38 
& \cellcolor[HTML]{f5fbf3} \color[HTML]{000000} 1.45
\\



GIN+ReverseMP \citep{jaume2019edgnnbidirecibmmaria}
& \cellcolor[HTML]{e1f3dc} \color[HTML]{000000} 14.29 
& \cellcolor[HTML]{a3da9d} \color[HTML]{000000} 36.96 
& \cellcolor[HTML]{c3e7bc} \color[HTML]{000000} 26.19 
& \cellcolor[HTML]{a2d99c} \color[HTML]{000000} 37.29 
& \cellcolor[HTML]{abdda5} \color[HTML]{000000} 34.29 
& \cellcolor[HTML]{bce4b5} \color[HTML]{000000} 28.57 
& \cellcolor[HTML]{eff9ec} \color[HTML]{000000} 5.58
\\
{\ +Ports} 
& \cellcolor[HTML]{a7dba0} \color[HTML]{000000} 35.71 
& \cellcolor[HTML]{208843} \color[HTML]{f1f1f1} 76.09 
& \cellcolor[HTML]{8ed08b} \color[HTML]{000000} 42.86 
& \cellcolor[HTML]{63bc6e} \color[HTML]{f1f1f1} 54.24 
& \cellcolor[HTML]{63bc6e} \color[HTML]{f1f1f1} 54.29 
& \cellcolor[HTML]{8ed08b} \color[HTML]{000000} 42.86 
& \cellcolor[HTML]{f2faf0} \color[HTML]{000000} 3.31
\\
{\ +EgoIDs (Multi-GIN)}
& \cellcolor[HTML]{8ed08b} \color[HTML]{000000} 42.86 
& \cellcolor[HTML]{5ab769} \color[HTML]{f1f1f1} 56.52 
& \cellcolor[HTML]{bce4b5} \color[HTML]{000000} 28.57 
& \cellcolor[HTML]{83cb82} \color[HTML]{000000} 45.76 
& \cellcolor[HTML]{6ec173} \color[HTML]{000000} 51.43 
& \cellcolor[HTML]{97d492} \color[HTML]{000000} 40.48 
& \cellcolor[HTML]{f6fcf4} \color[HTML]{000000} 0.83
\\

\midrule

Multi-GIN+EU
& \cellcolor[HTML]{8ed08b} \color[HTML]{000000} 42.86 
& \cellcolor[HTML]{077331} \color[HTML]{f1f1f1} 84.78 
& \cellcolor[HTML]{97d492} \color[HTML]{000000} 40.48 
& \cellcolor[HTML]{6abf71} \color[HTML]{000000} 52.54 
& \cellcolor[HTML]{2f974e} \color[HTML]{f1f1f1} 70.00 
& \cellcolor[HTML]{a7dba0} \color[HTML]{000000} 35.71
& \cellcolor[HTML]{f6fcf4} \color[HTML]{000000} 0.83
\\
Multi-PNA
& \cellcolor[HTML]{2c944c} \color[HTML]{f1f1f1} 71.43 
& \cellcolor[HTML]{005221} \color[HTML]{f1f1f1} 95.65 
& \cellcolor[HTML]{aedea7} \color[HTML]{000000} 33.33 
& \cellcolor[HTML]{2c944c} \color[HTML]{f1f1f1} 71.19 
& \cellcolor[HTML]{005622} \color[HTML]{f1f1f1} 94.29 
& \cellcolor[HTML]{3da65a} \color[HTML]{f1f1f1} 64.29 
& \cellcolor[HTML]{f6fcf4} \color[HTML]{000000} 0.62
\\

\bottomrule
\end{tabular}
    \endgroup
    \caption{Recall scores by money laundering pattern on AML Small LI dataset, where ground truth pattern labels are available.
    }
    \label{tab:aml_small_li_patterns}
\end{table*}

\subsection{Additional GNN Baselines}
\label{app:add_gnn_baselines}

There are many ``expressive'' GNN architectures that have been developed for simple (no loops or parallel edges), undirected graphs with subgraph detection in mind. For example, PPGN \citep{maron2019provably} and DropGIN \citep{papp2021dropgnn} to name a few. However, as we argue in \cref{sec:setup} \textit{Base GNNs and Baselines}, none of these has been developed for directed multigraphs, and their ability to detect subgraphs unfortunately does not translate into this setting. For instance, since none of these models consider the direction of edges, they cannot distinguish between incoming and outgoing neighbors. Some of these architectures could be adapted to the setting, and in some cases, our proposed adaptations offer a possible approach to do so. 
To support our argument, we have added more baselines for the subgraph detection experiments, please see \cref{tab:new_sim_results} for the results. When possible, we also include adapted versions based on our proposed adaptations. The architectures are indeed unable to perform well without adaptation.

\subsection{Complex Patterns and Dataset Size}
\label{app:complex_tasks}

Additional experiments were carried out to measure the influence of dataset size on the subgraph detection tasks. We restrict these experiments to the GIN model with all the adaptations. We also restrict to the more ``complex'' tasks that were not solved almost perfectly, namely C4, C5, C6, Scatter-Gather, and Biclique detection. Dataset sizes of 1024 up to 16384 are used, increasing by a factor of 2. All other graph generator settings are kept constant. In particular, we use the same average degree and average radius. 

The results can be seen in \cref{fig:sim_size_results}, and specific values can be found in \cref{tab:sim_size_results}. One can see that increasing the training dataset size gradually increases the scores on all the complex tasks, with even 6-cycle detection reaching above $97\%$ minority class F1 score. The starred results also show that restricting to a smaller set of tasks significantly improves performance.

For cycles, we further extended these experiments to cycles of length $10$ and training datasets with $65536$ nodes. These results can be seen in \cref{tab:sim_cycles} and show similar trends with all cycle sizes reaching above $98\%$ minority class F1 score with the largest training dataset.

\subsection{Subgraph Detection with Unique Node IDs}
\label{app:sim_nodes_ids}

We include additional subgraph detection results below. 
Firstly, we rerun the subgraph detection experiments with all GIN-based models with the addition of random unique node IDs as part of the input. The results can be seen in \cref{tab:sim_node_ids}. This experiment shows that theoretical expressiveness does not always translate into better results in practice. Comparing the first row of the table with the first row of \cref{tab:sim_results} reveals that random node IDs, although universal, are not effective for subgraph detection. The rest of the table shows the effect of adding adaptations. The trends are the same, and the results are similar to \cref{tab:sim_results}, but most of the scores are actually slightly lower with the node IDs.
Note in particular that both port numbers and ego IDs (in addition to reverse message passing) are still needed to attain high scores for the complex patterns. This highlights that although node IDs could easily provide either of these features, the neural network is not able to extract and use this information effectively. The results indicate that the adaptations offer the right inductive biases for the task.



\subsection{Subgraph Detection with the Collapsed Multigraph}
\label{app:sim_collapsed}

Secondly, we explore whether the directed multigraph input can be collapsed into a directed graph with weighted edges, where the edge weights indicate the number of parallel edges in the original multigraph. We rerun the subgraph detection experiments with all GIN-based models. The results of the collapsed multigraph experiments can be found in \cref{tab:sim_collapsed}. The results are very similar to using the 
But importantly, note that port numbers are no longer needed to solve the tasks. For example, fan-in can now be solved perfectly without ports, and port numbers do not increase any of the scores significantly, whether added on top of GIN or GIN with reverse message passing. This is not surprising as all of these tasks can be solved easily on a weighted directed graph (given reverse message passing and ego IDs as needed).
Indeed, collapsing the parallel edges of a multigraph in this way could be a valid option for some multigraph tasks. However, as soon as edges have non-additive features that are important for the task, this might no longer be an option. It might not be possible to summarize the features of parallel edges without losing valuable information.

\section{Detailed AML Results}
\label{app:detailed_aml_results}

Ground truth money laundering pattern labels are available for a portion of the fraudulent transactions in the AML datasets. The remaining illicit transactions (without pattern labels) do not belong to specific money laundering patterns; we might therefore expect worse recall scores for these transactions.  
Using these pattern labels, we calculate recall scores for each pattern type individually and present these results in \cref{tab:aml_small_hi_patterns} for AML Small HI and in \cref{tab:aml_small_li_patterns} for AML Small LI.

From the tables, we can see that our models identify most of the laundering patterns leading to very high individual recall scores, though the cycle and bipartite recall scores could be further improved. 
This is likely due to the fact that our best-performing GNN models used only two message passing layers, making it impossible to detect longer cycles. Increasing the number of layers can increase the overall F1 scores, but also comes at a significant cost in terms of the runtime.

We can also immediately notice that the recall scores for illicit transactions that do not belong to any specific money laundering pattern (none) are close to $0\%$. 
This probably explains the low aggregate minority class F1 scores for Small LI (\cref{tab:full_aml_eth_results}) where $71\%$ of the laundering transactions are not part of laundering patterns. The equivalent value for Small HI is only $38\%$.
These insights reveal just how effective the proposed approach is in detecting money laundering patterns.

}

\end{document}